\UseRawInputEncoding
\documentclass[10pt,journal,cspaper,compsoc]{IEEEtran}

\usepackage{times}
\usepackage{epsfig}
\usepackage{graphicx}
\usepackage{amsmath}
\usepackage{amssymb}
\usepackage{bm}
\usepackage{mathrsfs}
\usepackage{txfonts}
\usepackage{algorithm} 
\usepackage{enumerate}
\usepackage{multirow}

\usepackage{epstopdf}
\usepackage{setspace}

\ifCLASSOPTIONcompsoc
  \usepackage[nocompress]{cite}
\else
  \usepackage{cite}
\fi

\ifCLASSINFOpdf
\else
\fi

\usepackage[pagebackref=true,breaklinks=true,letterpaper=true,colorlinks,bookmarks=false]{hyperref}

\usepackage{color}
\definecolor{myColor}{rgb}{0, 0, 1}

\newcommand{\etal}{{\it et al. }}

\usepackage{multirow}

\usepackage[breaklinks=true,bookmarks=false]{hyperref}

\newcommand*{\vv}[1]{\vec{\mkern0mu#1}}

\usepackage{geometry}
\begin{document}

%
\title{HEMlets PoSh: Learning Part-Centric Heatmap Triplets for 3D Human Pose and Shape Estimation}
%
%
%
%

\author{Kun~Zhou,
        Xiaoguang~Han,~\IEEEmembership{Member,~IEEE,}
				Nianjuan~Jiang,~\IEEEmembership{Member,~IEEE,}
				Kui~Jia,~\IEEEmembership{Member,~IEEE,}
        and~Jiangbo~Lu,~\IEEEmembership{Senior~Member,~IEEE}
\IEEEcompsocitemizethanks{\IEEEcompsocthanksitem K.~Zhou, N.~Jiang and J.~Lu are with SmartMore Co., Ltd., Shenzhen, China. (Corresponding email:~jiangbo.lu@gmail.com)\protect
\IEEEcompsocthanksitem X.~Han is with Shenzhen Institute of Big Data, The Chinese University of Hong Kong (Shenzhen), Shenzhen, China.
\IEEEcompsocthanksitem K.~Jia are with South China University of Technology, Guangzhou, China.}}

\IEEEtitleabstractindextext{
\begin{abstract}
Estimating 3D human pose from a single image is a challenging task. This work attempts to address the uncertainty of lifting the detected 2D joints to the 3D space by introducing an intermediate state - Part-Centric Heatmap Triplets ({\emph{HEMlets}}), which shortens the gap between the 2D observation and the 3D interpretation. The HEMlets utilize three joint-heatmaps to represent the relative depth information of the end-joints for each skeletal body part. In our approach, a Convolutional Network~(ConvNet) is first trained to predict HEMlets from the input image, followed by a volumetric joint-heatmap regression. We leverage on the integral operation to extract the joint locations from the volumetric heatmaps, guaranteeing end-to-end learning. Despite the simplicity of the network design, the quantitative comparisons show a significant performance improvement over the best-of-grade methods (e.g. $20\%$ on Human3.6M). The proposed method naturally supports training with ``in-the-wild'' images, where only weakly-annotated relative depth information of skeletal joints is available. This further improves the generalization ability of our model, as validated by qualitative comparisons on outdoor images. Leveraging the strength of the HEMlets pose estimation, we further design and append a shallow yet effective network module to regress the SMPL parameters of the body pose and shape. We term the entire HEMlets-based human pose and shape recovery pipeline {\emph{HEMlets PoSh}}. Extensive quantitative and qualitative experiments on the existing human body recovery benchmarks justify the state-of-the-art results obtained with our HEMlets PoSh approach.
\end{abstract}

}

\maketitle

\IEEEdisplaynontitleabstractindextext

%
\IEEEpeerreviewmaketitle


\IEEEraisesectionheading{\section{Introduction}\label{sec:introduction}}
\IEEEPARstart{H}{uman} pose estimation from a single image is an important problem in computer vision, 
because of its wide applications, e.g., video surveillance and human-computer interaction. 
Given an image containing a single person, 3D human pose inference aims to predict 
3D coordinates of the human body joints. Recovering 3D information of human poses 
from a single image faces several challenges. The challenges are at least threefold: 1) reasoning about 3D human poses from a single image is by itself 
very difficult due to the inherent ambiguities; 2) for such a regression task, how to effectively bridge the gap between the 2D image and the target 3D human pose is actually challenging yet important; 3) for ``in-the-wild" images, both 3D capturing and manual labeling require 
a lot of effort to obtain high-quality 3D annotations, making the 
training data extremely scarce. 

For 2D human pose estimation, almost all best performing methods are detection 
based~\cite{SH-NET,ke2018multi,wei2016convolutional}. Detection-based approaches essentially 
divide the joint localization task into local image classification tasks. The latter 
is easier to train, because it effectively reduces the feature and target dimensions 
for the  learning system~\cite{sun2017integral}. 
Existing 3D pose estimation methods often use detection as an intermediate 
supervision mechanism as well. A straightforward strategy is to 
use volumetric heatmaps to represent the likelihood map of each 3D 
joint location~~\cite{pavlakos2017coarse}. Sun~{\it et al}.~\cite{sun2017integral} 
further proposed a differentiable soft-argmax operator that unifies the joint detection 
task and the regression task into an end-to-end training framework. 
This significantly improves the state-of-the-art 3D pose estimation accuracy. 

\begin{figure}
  \centering
  \includegraphics[width=\columnwidth]{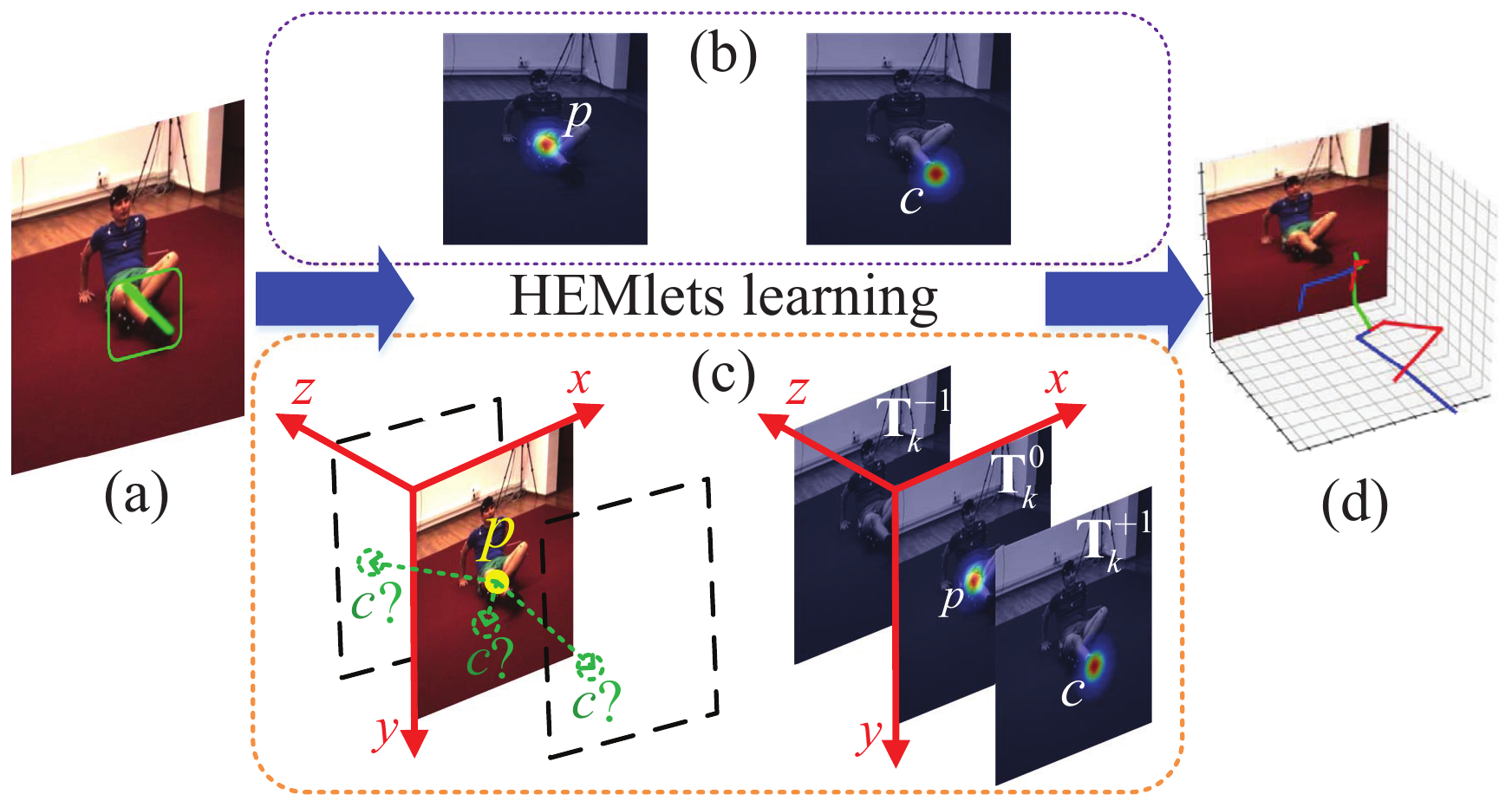}
	\caption{Overview of the HEMlets-based 3D pose estimation. (a)~Input RGB image. Our algorithm encodes (b)~the 2D locations for the joints $p$ and $c$, but also (c)~their relative depth relationship for each skeletal part $\vv{pc}$ into HEMlets. (d) Output 3D human pose.}
  \label{fig:teaser}
\end{figure}

In this work, we propose a novel effective intermediate representation for 3D pose estimation - 
\emph{Part-Centric Heatmap Triplets (HEMlets)} (as shown in Fig.~\ref{fig:teaser}). 
The key idea is to polarize the 3D volumetric space around each 
distinct skeletal part, which has the two end-joints kinematically connected. 
Different from 
\cite{pavlakos2018ordinal}, our relative depth information is represented as three polarized heatmaps, 
corresponding to the different state of the local depth ordering of the part-centric joint pairs. Intuitively, HEMlets encodes the co-location 
likelihoods of pairwise joints in a dense per-pixel manner with the coarsest 
discretization in the depth dimension. Instead of considering arbitrary joint pairs, we focus on kinematically connected 
ones as they possess semantic correspondence with the input image, and are thus a more effective target for the subsequent learning. 
In addition, the encoded relative depth information is strictly local for the part-centric joint pairs and suffers less from potential inconsistent data annotation. 

The proposed network architecture is shown in Fig.~\ref{fig:framework}. A ConvNet is first 
trained to learn the HEMlets and 2D joint heatmaps, which are then fed together with the high-level image features to another ConvNet to produce a volumetric 
heatmap for each joint. We leverage on the soft-argmax regression~\cite{sun2017integral} 
to obtain the final 3D coordinates of each joint. Significant improvements 
are achieved compared to the best competing methods quantitatively and qualitatively. 
Most notably, our HEMlets method achieves a MPJPE of 39.9mm on Human3.6M~\cite{ionescu2014human3}, yielding about $20\%$ improvement over one best-of-grade method~\cite{sun2017integral}. 

The merits of the proposed method lie in three aspects:
\begin{itemize}
  \item 
  \textbf{\textit{Learning strategy. }}
  Our method takes on a progressive learning strategy, and 
  decomposes a challenging 3D learning task into a sequence of easier sub-tasks with mixed 
  intermediate supervisions, i.e., 2D joint detection and HEMlets learning. HEMlets 
  is the key bridging and learnable component leading to 3D heatmaps, and is much easier to 
  train and less prone to over-fitting. Its training can also take advantage of existing labeled 
  datasets of relative depth ordering~\cite{pavlakos2018ordinal,shi2018fbi}. 
  \item
  \textbf{\textit{Representation power.}}~HEMlets is based on 2D per-joint heatmaps, but extends them by a couple of additional heatmaps 
  to encode local depth ordering in a dense per-pixel manner. It builds on top of 2D heatmaps 
  but unleashes the representation power, while still allowing leveraging the soft-argmax 
  regression~\cite{sun2017integral} for end-to-end learning. 
  \item 
  \textbf{\textit{Simple yet effective. }}
  The proposed method features a simple network architecture design, and it is easy to train and 
  implement. It achieves state-of-the-art 3D pose estimation results validated by the evaluations over all standard benchmarks. 
\end{itemize}

A preliminary version of this work on 3D human pose estimation was published in the IEEE/CVF Conference on Computer Vision (ICCV) 2019~\cite{hemletsiccv}. This paper makes a few major contributions and extensions over the initial conference version as follows. First, we extend the proposed HEMlets pose framework to further recover the human body model from the given input image. We design a simple body model regression network connected to the preceding HEMlets pose network to recover a SMPL human body mesh from a single color image. In addition, we also describe in detail the proposed method, the training process as well as the model architecture. Our code will be made publicly available at the project website. Second, we introduce and present a new weakly-annotated FBI dataset and elaborate its advantages in obtaining weak annotations for the relative depth relationship between a pair of skeletal joints. The comparison of the FBI dataset and the recent Ordinal dataset\cite{pavlakos2018ordinal} is also provided. Lastly, we conduct thorough experiments including more ablation studies, as well as quantitative and qualitative evaluations for the recovered human body shape and pose. Extensive experiments justify the state-of-the-art performance of the proposed HEMlets-based pose and shape estimation method (termed as {\it HEMlets PoSh}) on all mainstream benchmark datasets.


%
%
%
%

\section{Related Work}\label{sec:relatedWork}
In this section, we review the approaches that are based on deep ConvNets for 3D human pose estimation and 3D body model recovery from a single color image.

\subsection{3D Body Pose Estimation}
We first conduct the literature review of 3d pose estimation in the following four aspects.

\textbf{Direct Encoder-Decoder.} With the powerful feature extraction capability of deep ConvNets, many approaches~\cite{li20143d,tekin2016structured,park20163d} learn end-to-end 
\textit{Convolutional Neural Networks}~(CNNs) to infer human poses directly from the images. Li and Chen~\cite{li20143d} are 
the first who used CNNs to estimate 3D human pose via a multi-task framework. Tekin~{\it et al}.~\cite{tekin2016structured} designed an auto encoder to model the joint dependencies in a high-dimensional feature space. Park~{\it et al}.~\cite{park20163d} proposed fusing 2D joint locations with high-level image features to boost the estimation of 3D human pose. However, these single stage 
methods are limited by the availability of 3D human pose datasets and cannot take advantage of large-scale 2D pose datasets that are vastly available.

\textbf{Transition with 2D Joints.} To avoid collecting 2D-3D paired data, a large number of works~\cite{ronchi2018s,zhou2017towards,yang20183d,martinez2017simple,fang2018learning,shi2018fbi} decompose the task of 3D pose estimation into two independent stages by: 1) firstly inferring 2D joint locations using well-studied 2D pose estimation methods, such as~\cite{zhou2017towards,ronchi2018s}; 2) and then learning a mapping to lift them into the 3D space. These approaches mainly focus on tackling
the second problem. For example, a simple fully connected residual network is proposed by Martinez~{\it et al}.~\cite{martinez2017simple} to directly recover 3D human pose from its 2D projection. Fang~{\it et al}.~\cite{fang2018learning} considered prior knowledge of human body configurations and proposed human pose grammar, leading to better recovery of the 3D pose from only 2D joint locations. Yang~{\it et al}.~\cite{yang20183d} adopted an adversarial learning scheme to ensure the anthropometrical validity of the output pose and further improved the performance. Recently, by involving a reprojection mechanism, the proposed method in~\cite{wandt2019repnet} shows insensitivity to overfitting and 
accurately predicts the result from noisy 2D poses. Though promising results have been achieved by these two-stage methods, 
a large gap exists between the 3D human pose and its 2D projections due to inherent ambiguities.

\textbf{3D-Aware Intermediate States.} To further bridge the gap between the 2D image and the target 3D human pose under estimation, some recent works~\cite{pavlakos2017coarse,shi2018fbi,pavlakos2018ordinal,sun2017integral} proposed to involve 3D-aware states for intermediate supervisions. Namely, a network is firstly trained to map the input image to these 3D-aware states, and then another network is trained to convert those states to the 3D joint locations. Finally, these two networks are combined and optimized jointly. A volumetric representation for 3D joint-heatmaps is proposed in ~\cite{pavlakos2017coarse}, with which the 3D pose is regressed in a coarse-to-fine manner. However, regressing a probability grid in the 3D space globally is also a very challenging task. It usually suffers from quantization errors for the joint locations. To address this issue, Sun~{\it et al}.~\cite{sun2017integral} exploited a soft-argmax operation and proposed 
an end-to-end training scheme for the 3D volumetric regression, 
achieving by far the best performance on 3D pose estimation. 
Inspired by~\cite{pons2014posebits} that the relative depth ordering 
across joints is helpful for resolving pose ambiguities, Pavlakos~{\it et al}.~\cite{pavlakos2018ordinal} adopted a ranking loss for pairwise ordinal depth to train the 3D human pose predictor explicitly. A similar scheme of relative depth supervision is utilized in the work of~\cite{ronchi2018s}. Forward-or-Backward Information~(FBI), proposed in ~\cite{shi2018fbi}, is another kind of relative depth information but focuses more on the bone orientations. Recently, Sharma~{\it et al}.~\cite{Sharma_2019_ICCV} proposed to train a deep conditional variational autoencoder to map 2D poses to 3D poses by learning ordinal maps. In this work, we propose HEMlets, a novel representation that encodes both 
2D joint locations and the part-centric relative depth ordering simultaneously. Experiments justify that this representation reaches by far the best balance between representation efficiency and learning effectiveness. 

\textbf{``In-the-Wild'' Adaptation.}
All the aforementioned approaches are mainly trained on the datasets collected under indoor settings, due to the difficulty of annotating 3D joints for ``in-the-wild'' images ~\cite{bourdev2009poselets}. 
Thus, many strategies are developed to make domain adaptation. 
By exploiting graphics techniques, previous works~\cite{varol2017learning,chen2016synthesizing} have synthesized a large ``faked'' dataset mimicking real images. Though these data benefit 3D pose estimation, they are still far from realistic, making the applicability limited. Recently, both Pavlakos~{\it et al}.~\cite{pavlakos2018ordinal} and Shi~{\it et al}.~\cite{shi2018fbi}  proposed to label the relative depth relationship across joints 
instead of the exact 3D joint coordinates. This weak annotation scheme 
not only makes building large-scale ``in-the-wild'' datasets 
feasible but also provides 3D-aware information 
for training the inference model in a weakly-supervised manner. 
With HEMlets representation, we can readily use these weakly annotated ``in-the-wild'' data for domain adaptation.

\subsection{3D Body Model Recovery}

In the recent years, 3D full body models have become popular, which are typically represented with a parametric human body space, such as SMPL~\cite{SMPL:2015}. The advantage is that a human body mesh can be easily generated from a set of body shape and pose parameters, so this turns the task of recovering a 3D body model from a single image into a problem of solving for a set of parameters. As a pioneering work in this arena, a two-stage framework~\cite{bogo2016keep} is proposed. It firstly infers 2D skeleton joints from the input image with a CNN-based model, and then searches the optimal parameters of SMPL to fit the joints with an optimization approach. Due to the depth ambiguity, the second stage tends to converge to a local minimum. To better address this problem, many works~\cite{kanazawa2018end,pavlakos2018learning,kolotouros2019convolutional} proposed to build an end-to-end pipeline to map images into the parametric space using deep regression models. 

However, the key challenge to train regression-based models is the lack of paired data, due to the inherent difficulty of annotating or capturing a groundtruth 3D model for a person instance. Existing approaches address this challenge roughly along three main directions. First, some works directly tackled the issue by putting efforts on constructing target datasets. The work of~\cite{varol2017learning} firstly built a synthetic dataset using graphics techniques, however, training only on this dataset is still difficult to produce a model that is applicable to real images. Lassner~{\it et al}.~\cite{lassner2017unite} proposed to apply the algorithm of~\cite{bogo2016keep} to obtain 3D body models for real images and then manually sift out the reasonable results, to build the final human body dataset. Unfortunately, the obtained 3D human shapes are still non-ideal and contain erroneous body part results. 

For the second category, several works attempted to directly add extra constraints on the output parameters to ease the training process. For example, the strategy of adversarial learning was utilized in~\cite{kanazawa2018end}, where a discriminator is exploited to constrain the regressed parameters against a reasonable distribution. Given the regressed SMPL models, Kolotouros~{\it et al}.~\cite{kolotouros2019spin} further adopted the approach of~\cite{bogo2016keep} to obtain better parameters to fit the input images. They then directly set an extra objective for the regression model to enforce its output parameters equal to the optimized ones. 

Lately, a large body of works proposed to use 2D intermediate representations, such as silhouettes~\cite{tan2018indirect,pavlakos2018learning} and densepose~\cite{alp2018densepose}-based representation~\cite{xu2019denserac,guler2019holopose}, which leverage on the idea of self-supervised training. Specifically, other than the main branch of mapping input images to SMPL meshes, these methods also constructed a novel branch to convert the input images to the proposed intermediate representations. A differentiable mesh renderer was then utilized to render the output meshes, which are compared with the learnt intermediate representations. This brings extra supervisions to guide the training process. 

More recently, some other 3D representations, such as voxels, mesh and UV-maps, have been used for building generative neural networks to infer 3D models from images. The work of~\cite{varol2018bodynet}, as the first attempt of this kind, proposed an approach to generate a voxel representation of a 3D body from a single image using 3D ConvNets. By taking a template human body mesh as an extra input and treating the mesh as a graph representation, Kolotouros~{\it et al}.~\cite{kolotouros2019convolutional} trained a Graph ConvNet to learn the deformation of the template model for fitting both the target pose and shape. The algorithm of DenseBody~\cite{yao2019densebody} represented the 3D body model with a parameterized UV-map, and then turned the task of geometry inference into a problem of image synthesis. This method further advances body mesh reconstruction accuracy. 

In this work, we find that the impact of 3D pose estimation on the accuracy of the final recovered body model is much more significant than the regressed shape parameters. Based on the estimated 3D pose obtained with our HEMlets pose approach, we show that a simple body regression method for SMPL model inference outperforms all the afore-discussed approaches.

\begin{figure}[t]
\centering 
\includegraphics[width=\linewidth]{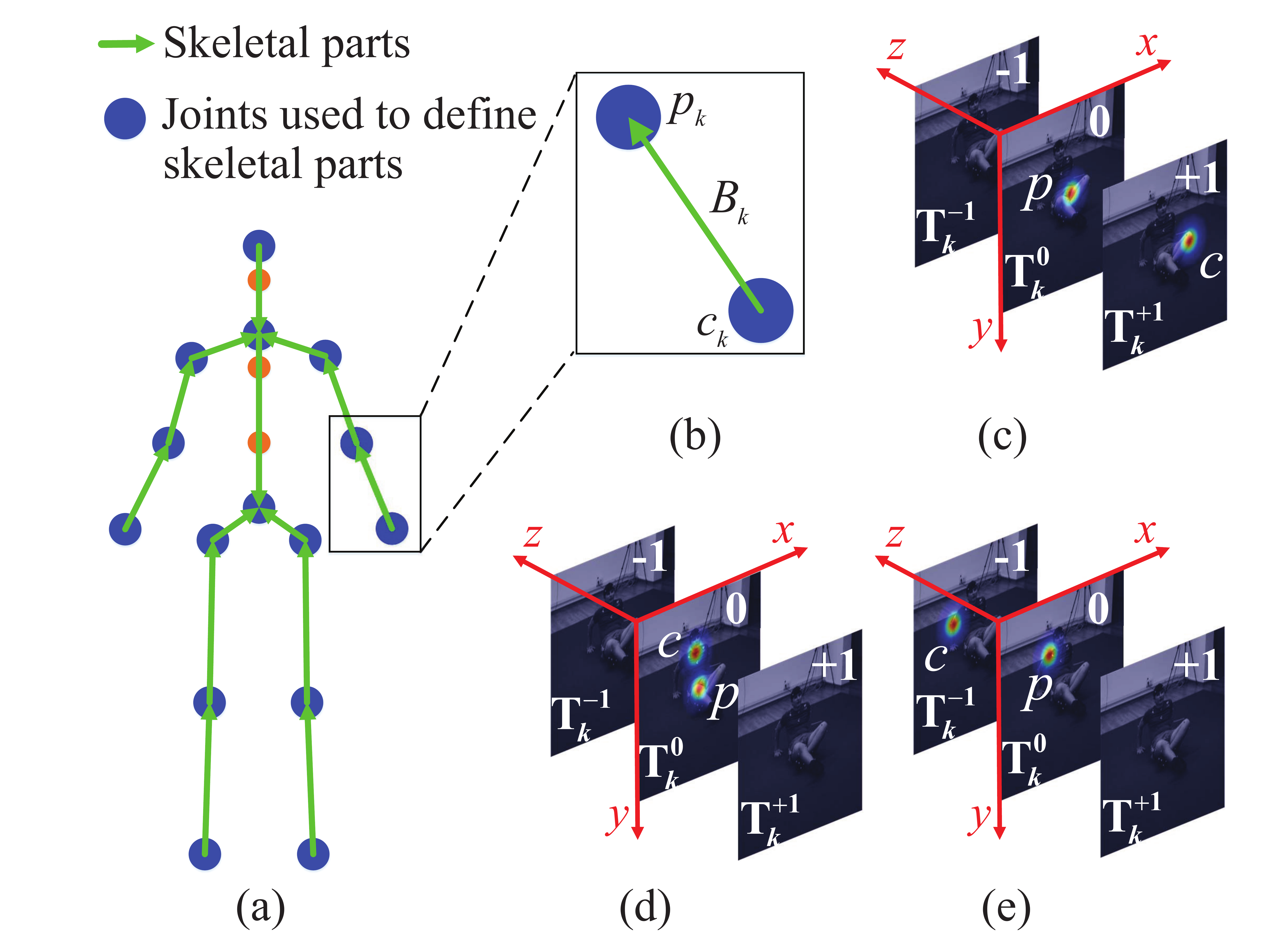}
\caption{Part-centric heatmap triplets $\{{\bf T}^{-1}_k,{\bf T}^{0}_k,{\bf T}^{+1}_k\}$ where $p$ and $c$ are the parent joint and the child joint. (a,~b) Joints and skeletal parts. We locate the parent joint $p$ of the $k$-th skeletal part $B_k$ at the zero polarity heatmap ${\bf T}^{0}_k$~(c-e). The child joint $c$ is located, according to relative depth of $p$ and $c$, in the positive~(c), zero~(d) and negative polarity heatmap~(e), respectively. }
\label{fig:tri_heatmaps}
\end{figure}

\begin{figure*}[t]
\centering
\includegraphics[width=16.5cm]{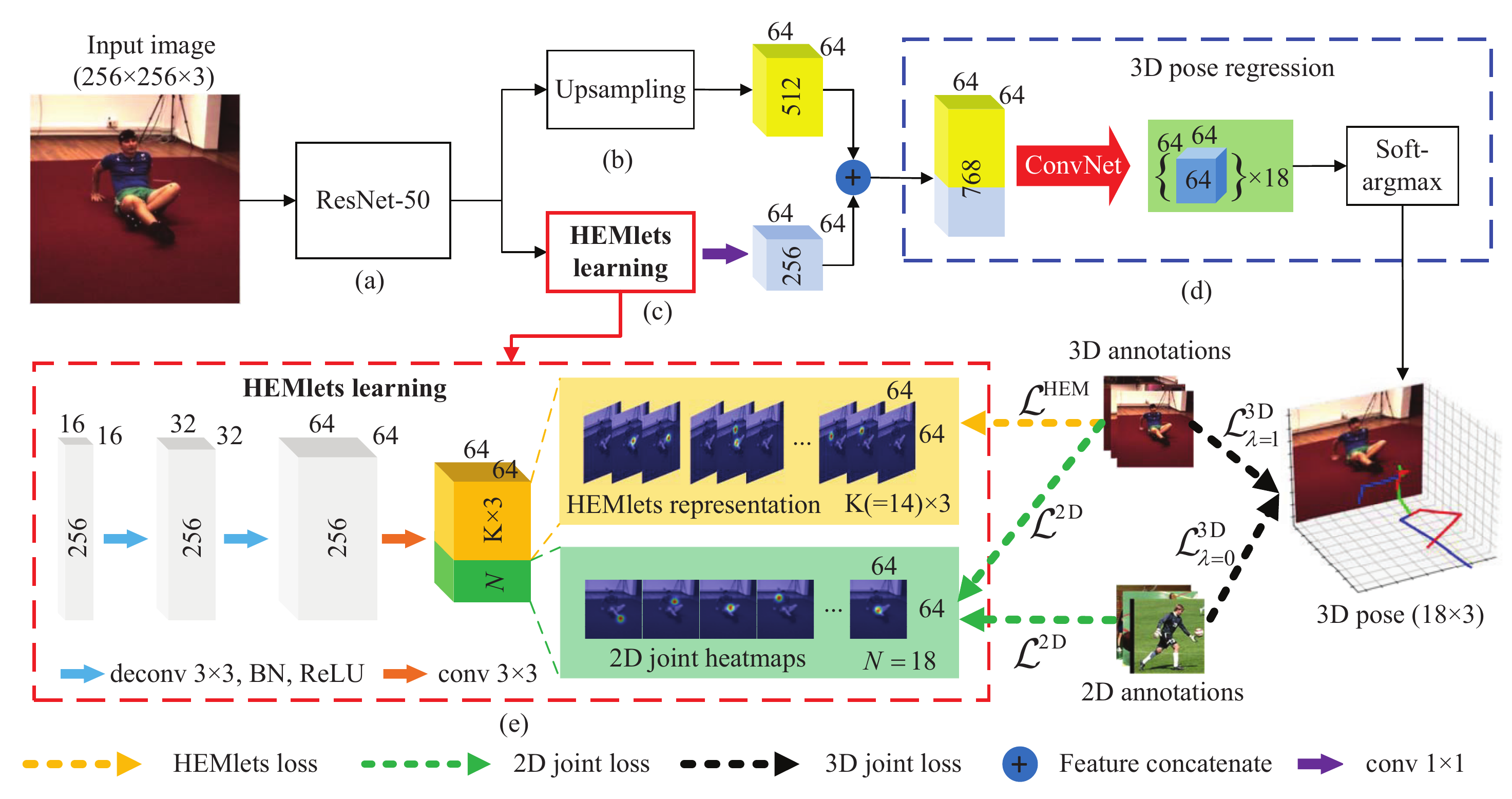}
\caption{The network architecture of our proposed approach. It consists of 
four major modules: (a) A ResNet-50 backbone for image feature extraction. (b) A ConvNet for image feature upsampling. (c) Another ConvNet for HEMlets learning and 2D joint detection. 
(d) A 3D pose regression module adopting a soft-argmax operation for 3D human pose estimation. (e) Details of the HEMlets learning module. ``Feature concatenate" denotes concatenating the feature maps from the HEMlets learning branch and the upsampling branch together.}
\label{fig:framework}
\end{figure*}

\section{HEMlets Pose Estimation} \label{sec:method}
We propose a unified representation of heatmap triplets to model the local information of body skeletal parts, i.e., kinematically connected joints, whereas the corresponding 2D image coordinates and relative depth ordering are considered. By such a representation, images annotated with relative depth ordering of skeletal parts can be treated equally with images annotated with 3D joint information. While the latter is usually very scarce, the former is relatively easy to obtain~\cite{shi2018fbi,pavlakos2018ordinal}. In this section, we first present the proposed part-centric heatmap triplets and its encoding scheme. Then, we elaborate a simple network architecture that utilizes the part-centric heatmap triplets for 3D human pose estimation.


\subsection{Part-Centric Heatmap Triplets}
\label{sec:constructHemlets}

We divide the full body skeleton consisting of $N=18$ joints into $K=14$ parts as shown in Fig.~\ref{fig:tri_heatmaps}(a).
Specifically, we use $B$ to denote the set of skeletal parts,
where $B = \left\{ B_1,B_2,\ldots,B_K \right\}$. For each part,
we denote the two associated joints as $(p, c)$, with $p$ being the parent node and $c$ being the child node. The relative depth ordering, denoted as $r(z_p, z_c)$, can be then described as a tri-state function~\cite{pavlakos2018ordinal,shi2018fbi}:
\begin{equation} 
r(z_p,z_c) =
\begin{cases}
1& \text{$z_p - z_c>\epsilon$}\\
0& \text{$\left| z_p - z_c \right|<\epsilon$} \\
-1& \text{$z_p - z_c<-\epsilon$}
\end{cases} ,
\label{eq:polarity}
\end{equation}

\noindent where $\epsilon$ is used to adjust the sensitivity of the function to the relative depth difference. The absolute depths of the two joints $p$ and $c$  are denoted by $z_p$ and $z_c$, respectively.

We argue that directly using the discretized label as an intermediate state for learning the
3D pose from a 2D joint heatmap, as was done in \cite{pavlakos2018ordinal,shi2018fbi}, is not
as effective. Since this abstraction tends to lose some important features encoded in 
the joints' spatial domain. Instead of elevating the
problem straight away to the 3D volumetric space, we utilize an
intermediate representation of the 3D-aware relationship of the parent joint $p_k$
and the child joint $c_k$ of a skeletal part $B_k$. Provided with the supervision signals, we define
polarized target heatmaps where a pair of normalized Gaussian peeks
corresponding to the 2D joint locations are
placed accordingly across three heatmaps (see Figure~\ref{fig:tri_heatmaps}).
We term them as the \textit{negative polarity heatmap} ${\bf T}^{-1}_k$, the \textit{zero polarity heatmap} ${\bf T}^{0}_k$ and the
\textit{positive polarity heatmap} ${\bf T}^{+1}_k$ with respect to the function value in
Eq.~(\ref{eq:polarity}). The parent joint $p_k$ is always placed in the
zero polarity heatmap ${\bf T}^{0}_k$. The child joint $c_k$ will appear in the negative/positive
polarity heatmap, if its depth is larger/smaller than that of the parent joint $p_k$~(i.e., $|r(z_p,z_c)| \neq 0$).
Both parent and child joints are co-located in the zero polarity heatmap if their
depths are roughly the same~(i.e., $r(z_p,z_c)=0$).

Formally, we denote the heatmap triplets of the skeletal part $B_k$ as the stacking
of three heatmaps ${\bf T}^{-1}_k,{\bf T}^{0}_k,{\bf T}^{+1}_k$: 
\begin{equation}
{\bf T}_k = \operatorname{Stack}[{\bf T}^{-1}_k,{\bf T}^{0}_k,{\bf T}^{+1}_k] ,
\end{equation}

Given 3D groundtruth coordinates of all joints, 
we can readily compute the heatmap triplets of each skeletal part. For 
easy reference, we shall refer to the part-centric heatmap triplets 
${\bf T}_k$ as \textit{HEMlets}, and use it afterwards.\\  


{\bf Discussions.}
Here we provide some understandings of HEMlets from a few perspectives. First, different from a joint-specific 2D heatmap that models the detection likelihood for each intended joint on the $(x, y)$ plane, HEMlets models part-centric pairwise joints' co-location likelihoods on the $(x, y)$ plane simultaneously with their ordinal depth relations. This helps to learn geometric constraints (e.g., bone lengths) implicitly. Second, by augmenting a 2D heatmap to a triplet of heatmaps, HEMlets learns and evaluates the co-location likelihood for a pair of connected joints $(p,c)$ by the joint probability distribution $P(x_p,y_p,x_c,y_c,r(z_p,z_c))$ in a locally-defined volumetric space. In contrast, Pavlakos~{\it et al}.~\cite{pavlakos2018ordinal} relaxed the learning target and marginalized the 3D probability distributions independently for the $(x, y)$ plane i.e., $P(x_p,y_p), P(x_c,y_c)$ and the $z$-dimension, with the latter supervised independently by $r(z_p,z_c)$ based on a ranking loss. Third, by exploiting the available supervision signals to a larger extent, HEMlets brings the benefit of making the knowledge more explicitly expressed and easier to learn, and bridges the gap in learning the 3D information from a given 2D image.


\subsection{3D Pose Inference}
\noindent 
{\bf Network architecture.}
We employ a fully convolutional network to predict the 3D human pose as 
illustrated in Figure~\ref{fig:framework}. A ResNet-50~\cite{ghiasi2014occlusion} 
backbone architecture is adopted for basic feature extraction. 
One of the two upsampling branches is used to learn the HEMlets 
and the 2D heatmaps of skeletal joints, and the other one 
is used to perform upsampling of the learned features to the same resolution 
as the output heatmaps. Both HEMlets and the 2D joint heatmaps are then 
encoded jointly by a 2D convolutional operation to form a latent global 
representation. Finally these global features are joined with the convolutional features 
extracted from the original image to predict a 3D feature map for each joint. 
We perform a soft-argmax operation~\cite{sun2017integral} to aggregate 
information in the 3D feature maps to obtain the 3D joint estimations. \\

\label{sec:training}
{\bf HEMlets loss.}
Let us denote with ${\bf T}^{\rm gt}$ the groundtruth HEMlets of all skeletal parts 
and with ${\bf \hat{T}}$ the corresponding prediction. We use a standard $L_2$ 
distance between ${\bf T}^{\rm gt}$ and ${\bf \hat{T}}$ to compute the HEMlets loss as follows:
\begin{equation}
{\mathcal{L}}^{\rm HEM} = {\lVert ({\bf T}^{\rm gt} - {\bf \hat{T}}) \odot {\bf \Lambda} \rVert}_2^2 ,
\label{eq:tri-state-loss}
\end{equation}
where $\odot$ denotes an element-wise multiplication, and ${\bf \Lambda}$ is a binary tensor 
to mask out missing annotations.\\

{\bf Auxiliary 2D joint loss.}
As HEMlets essentially contains heatmap responses of 2D joint locations, we 
adopt a heatmap-based 2D joint detection scheme to facilitate HEMlets prediction. 
The $L_2$ loss of 2D joint prediction is computed as:
\begin{equation}
\mathcal{L}^{\rm 2D} = \sum_{n=1}^N{\lVert {\bf H}^{\rm gt}_n -  {\bf \hat{H}}_n \rVert}_2^2,
\label{eq:heat-loss}
\end{equation}
where ${\bf H}^{\rm gt}_n$ is the groundtruth 2D heatmap of the $n$-th 2D joint 
and ${{\bf{\hat{H}}}_n}$ is the corresponding network prediction.\\

{\bf Soft-argmax 3D joint loss.}
To avoid quantization errors and allow end-to-end learning, Sun~{\it et al}.~\cite{sun2017integral} 
suggested soft-argmax regression for 3D human pose estimation. Given learned
volumetric features ${\bf F}_n$ of size $(h \times w \times d)$ for the $n$-th joint,
the predicted 3D coordinates are given as:

\begin{equation}
[\hat{x}_n,\hat{y}_n,\hat{z}_n] = \int_{\mathbf{v}}{ \mathbf{v} \cdot \operatorname{Softmax}({\bf F}_n)},
\label{eq:Integral}
\end{equation}
where $\mathbf{v}$ denotes a voxel in the volumetric feature space of ${\bf F}_n$. For robustness, we employ the $L_1$ loss for the regression of 3D joints. Specifically, the loss is defined as:

\begin{equation}
\mathcal{L}^{\rm 3D}_{\lambda} = \sum_{n=1}^N{ ( \left| x_n^{\rm gt} -  {\hat{x}}_n\right| + \left| y_n^{\rm gt} -  {\hat{y}}_n\right| + \lambda \left| z_n^{\rm gt} -  {\hat{z}}_n\right| )},
\label{eq:3d-loss-lambda}
\end{equation}
where the groundtruth 3D position of the  $n$-th joint is given as $(x_n^{\rm gt},y_n^{\rm gt},z_n^{\rm gt})$. We use the same 2D and 3D mixed training strategy in~\cite{sun2017integral}~($\lambda \in \{0,1\}$): $\lambda$ in Eq.~(\ref{eq:3d-loss-lambda}) is set to $1$ when the training data is from 3D datasets, and $\lambda = 0$ when the data is from 2D datasets. \\

{\bf Training strategy.}
For HEMlets prediction, We combine $\mathcal{L}^{\rm HEM}$ and $\mathcal{L}^{\rm 2D}$ for the intermediate supervision. 
The loss function is defined as:
\begin{equation}
\mathcal{L}^{\rm int} =  \mathcal{L}^{\rm HEM} + \mathcal{L}^{\rm 2D}.
\label{eq:3d-loss}
\end{equation}
By using $\mathcal{L}^{\rm HEM}$ and $\mathcal{L}^{\rm 2D}$ jointly as supervisions, we allow training the 
network using images with 2D joint annotations and 3D joint annotations. By 
3D joint annotation, we refer to annotations with exact 3D joint coordinates or 
relative depth ordering between part-centric joint pairs. 

The end-to-end training loss $\mathcal{L}^{\rm tot}$ is defined by combining $\mathcal{L}^{\rm int}$ with $\mathcal{L}^{\rm 3D}_{\lambda}$:
\begin{equation}
\mathcal{L}^{\rm tot} = \alpha * \mathcal{L}^{\rm int}  + \mathcal{L}^{\rm 3D}_{\lambda},
\label{eq:mix-loss}
\end{equation}
where $\alpha=0.05$ in all our experiments.\\

\subsection {\bf Implementation Details}

\begin{figure}
  \centering
  \includegraphics[width=0.9\columnwidth]{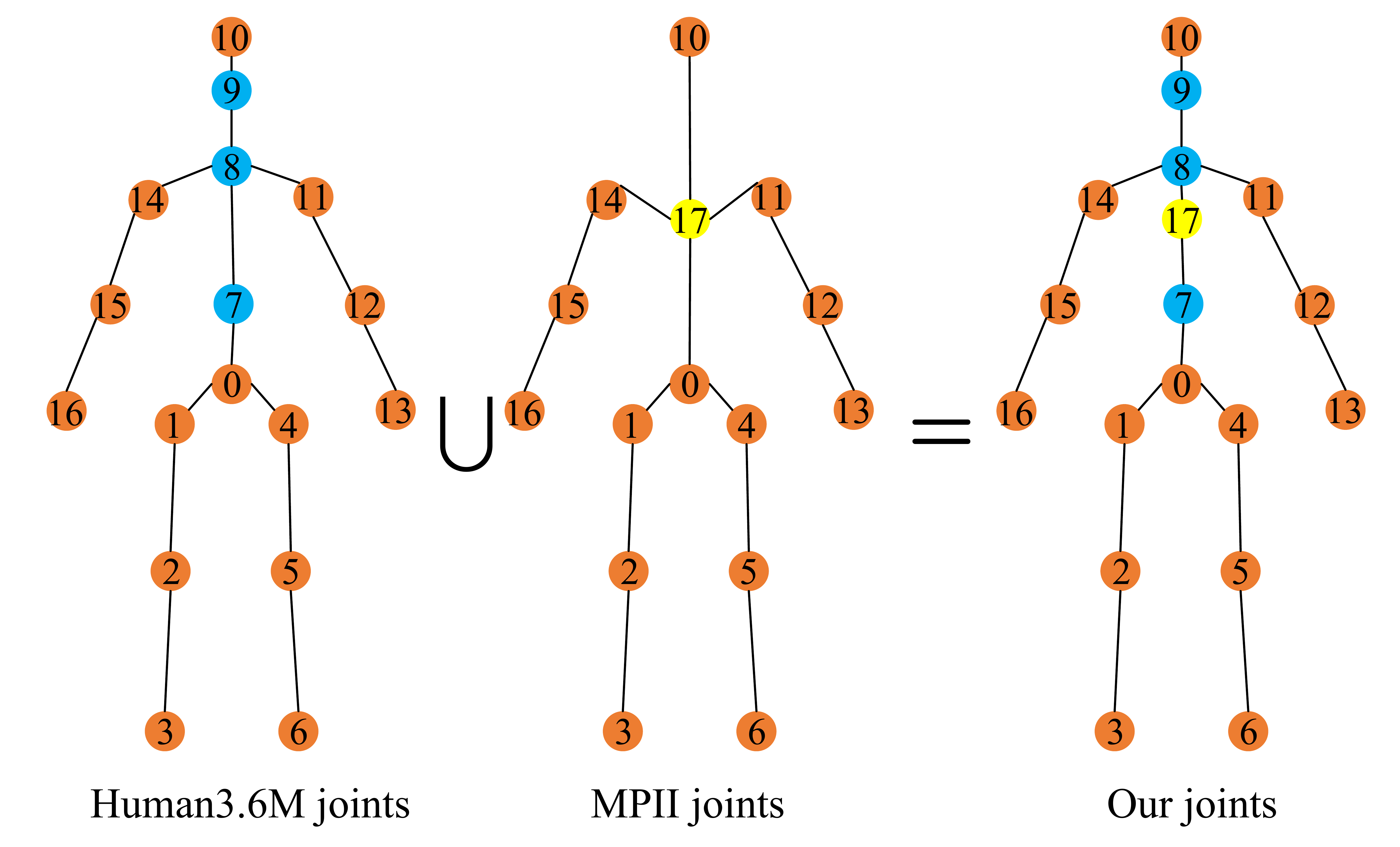}
  \caption{A unified body joint definition adopted in our method by merging the joints defined by the Human3.6M and MPII datasets.}
  \label{fig:JointTraining}
\end{figure}

Now we present a few implementation details in the proposed method. As different human pose datasets may have different definitions for body joints, we choose to accommodate this difference from different supervision sources. The purpose is to take advantage of more human pose annotation sources, when using the 2D and 3D mixed training strategy~\cite{sun2017integral}. Figure~\ref{fig:JointTraining} illustrates the joint structures defined by the Human3.6M~\cite{ionescu2014human3} and MPII~\cite{Andriluka} datasets, as well as our joint structure definition. We take the union of these two sets of joint definitions to form a 18-joint set as the regression target. Suppose performance evaluation is conducted on the Human3.6M dataset, then only those estimated joints used by Human3.6M will be evaluated, as in Martinez~\etal's work~\cite{martinez2017simple}.

To prepare the human bounding-box input for the proposed network, we crop from the original input image a square-shaped region based on the ground-truth bounding-box, and then resize it proportionally to $256 \!\times\! 256$. To obtain the final metric scale prediction from the network output (in voxel/pixel space), we resort to the average body bone length learned during the training phase to enable this prediction mapping. We did not use the ground-truth (depth) information during the test phase, e.g. the distance to the root/pelvis for obtaining the scaling factor.

We implement our method in PyTorch. The model is trained in an end-to-end manner 
using both images with 3D annotations (e.g., Human3.6M~\cite{ionescu2014human3} or   
HumanEva-I~\cite{sigal2010humaneva}), and 2D annotations~(MPII~\cite{Andriluka}). 
In our experiments, we adopt an adaptive value of $\epsilon$ in 
Eq.~(\ref{eq:polarity}) for each skeletal part: 
${\epsilon}_k = 0.5 {\lVert B_k \rVert}$~($ {\lVert B_k \rVert}$ is the 
3D Euclidean distance between the two end joints of the skeletal part $B_k$).
The training data is further augmented with rotation ($\pm30^{\circ}$), 
scale ($0.75\!-\!1.25$), horizontal flipping (with a probability of $0.5$) and color distortions.  
By using a batch size of $64$, a learning rate of $0.001$ and Adam optimization, the 
training took $100$K iterations to converge. It took about a few days~($2\!-\!4$) with 
four NVIDIA GTX 1080 GPUs to train the HEMlets pose estimation model.

\section{HEMlets Body Model Regression} \label{sec:shape_method}

So far, we have presented the proposed HEMlets pose estimation method in detail. It is natural to consider whether the proposed method can be extended also to recover human body models from the given input color images. To this end, we design and append a shallow yet effective network module to the preceding HEMlets pose network, which leverages the 3D pose estimation accuracy to regress the parameters of the body shape and pose. In this work, we employ the popular 3D body SMPL model~\cite{SMPL:2015}, where a human body mesh is parameterized by a 3D body shape parameter $\beta \in \mathbb{R}^{10}$ and a pose parameter $\theta \in \mathbb{R}^{24 \times 3}$.

\begin{figure}[t]
\centering 
\includegraphics[width=\linewidth]{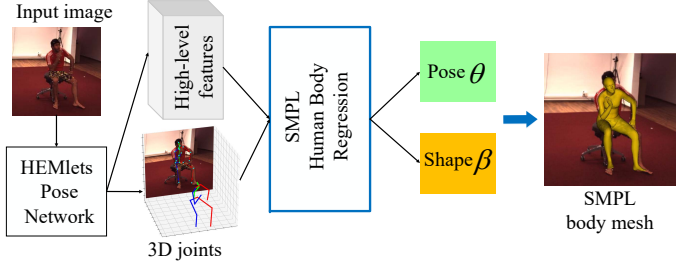}
\caption{HEMlets-based parametric 3D human body regression from a single color image. We append a shallow yet effective SMPL body mesh regression network to the preceding HEMlets pose estimation network, which is trained end-to-end to regress the SMPL shape and pose parameters $\{\beta, \theta\}$.}
\label{fig:shapeModule}
\end{figure}

As shown in Fig.~\ref{fig:shapeModule}, the newly added body model regression module is very simple. It takes the predicted 3D joint coordinates from the early stage as input, together with the high-level image features extracted from the given color image. This regression module is trained to regress the SMPL shape and pose parameters as final outputs. It is worth noting that we do not perform explicit human image segmentation, but instead use the high-level image features as implicit cues for shape regression.

In our implementation, the additional regression module is trained together with the 3D pose network in an end-to-end manner. Similar to recent works~\cite{xu2019denserac,kolotouros2019convolutional}, the SMPL pose parameter $\theta$ is converted into 24 rotation matrices for pose regression, which avoids the known singularity problem of the axis-angle representation. Following the similar strategy, the SMPL pose loss $\mathcal{L}_{\theta}$ is defined as: 
\begin{equation}
\mathcal{L}_{\theta} = \sum_{i=1}^{24} {\lVert R_i^{\rm gt} - \hat{R_i} \rVert} \;,
\label{eq:pose loss}
\end{equation} 
where $R_i$ denotes the rotation matrix corresponding to the $i$-th joint. The SMPL shape regression loss $\mathcal{L}_{\beta}$ is simply computed using the $L_1$ loss as,
\begin{equation}
\mathcal{L}_{\beta} = \sum_{i=1}^{10} {\lVert {\beta}_i^{\rm gt} - \hat{{\beta}_i} \rVert} \;.
\label{eq:beata loss}
\end{equation}

Finally, the end-to-end training loss ${L}_{\rm mesh}$ for the parametric 3D human body regression is given by
\begin{equation}
\mathcal{L}_{\rm mesh} = \mathcal{L}_{\theta}  + \mathcal{L}_{\beta} + \mathcal{L}^{\rm tot} \;,
\label{eq:shape loss}
\end{equation} 
where $\mathcal{L}^{\rm tot}$ is the total loss defined for pose estimation in Eq.~(\ref{eq:mix-loss}).


\vspace{-2pt}
\section{Weakly-Annotated FBI Dataset} \label{sec:fbi}

{
\begin{figure}
  \centering
  \includegraphics[width=\columnwidth]{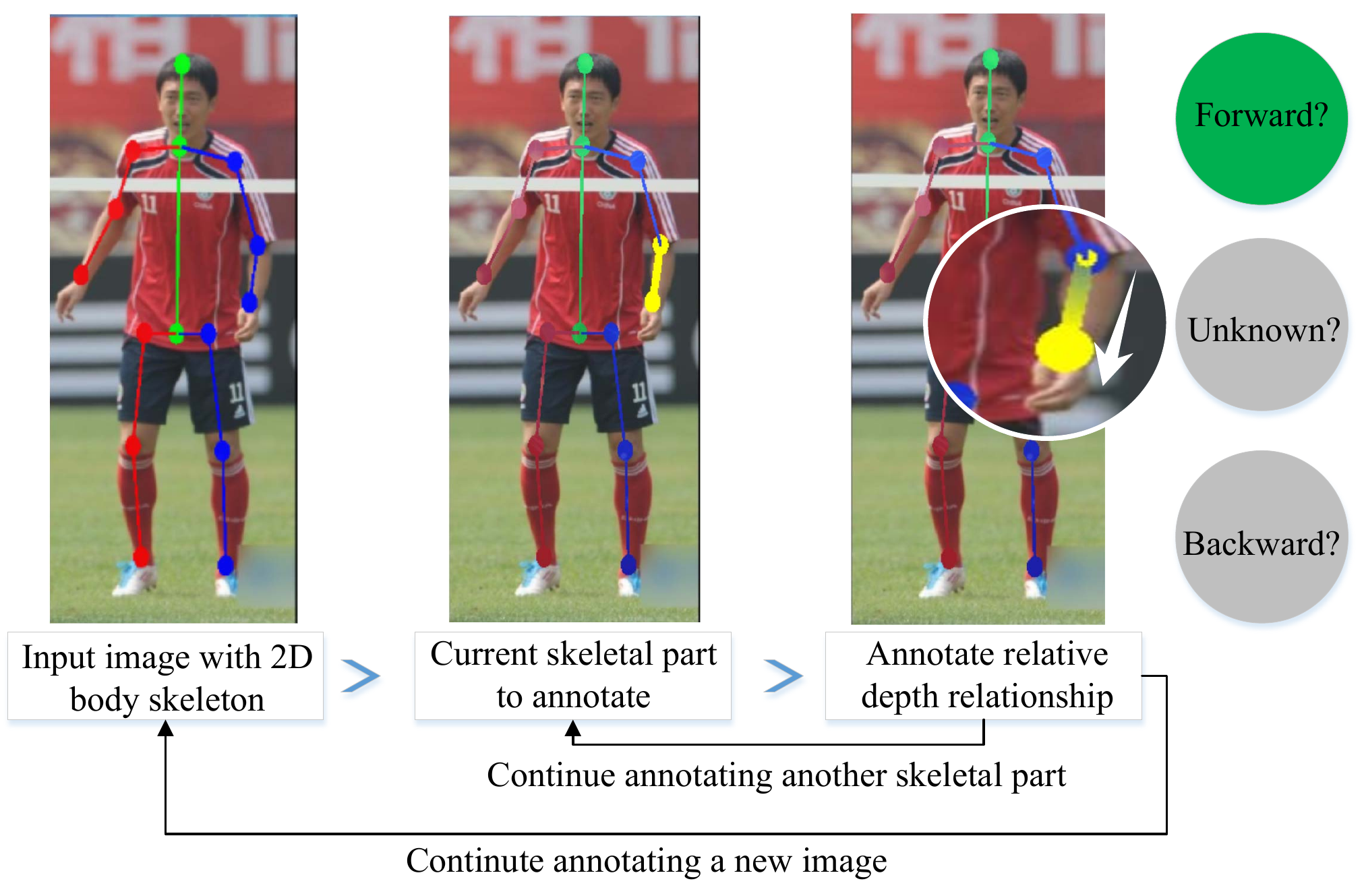}
  \caption{User annotation interface for obtaining the weakly-annotated FBI dataset. An annotator is asked to assign a label of either ``\textit{Backward}", ``\textit{Forward}" or ``\textit{Unknown}" to a given skeletal part.}
  \label{fig:annotation}
\end{figure}

In this section, we introduce a new Forward-or-Backward Information (FBI) dataset, and elaborate its advantages in obtaining weak annotations for the relative depth relationship between a pair of skeletal joints. To prepare this FBI dataset, 12K images are randomly drawn from the MPII dataset~\cite{Andriluka}, for which only 2D joint annotations are available. Then, each body part is assigned with a label of either ``\textit{Backward}", ``\textit{Forward}" or ``\textit{Unknown}". We designed a simple user interface to facilitate the annotation. As shown in Fig.~\ref{fig:annotation}, an annotator was presented with one image at a time with the native 2D skeleton overlaid over the input image. The annotator was asked to assign ``\textit{Backward}" or ``\textit{Forward}" labels to only a subset of the body parts for which she/he is confident with. The rest of the body parts are assigned with the ``\textit{Unknown}" labels by default.

\subsection{Comparison with The Ordinal Dataset~\cite{pavlakos2018ordinal}}

At the first glance, both FBI and Ordinal~\cite{pavlakos2018ordinal} annotation schemes aim at annotating the depth ordering between two body joints. However, the FBI scheme simplifies the annotation objective and reduces the annotation complexity with a good reason. The Ordinal scheme tries to annotate the relative depth information between every pair of joints. For each image, there are $\binom{14}{2}=91$ questions that need to be answered by the annotator. This annotation requirement is not only time-consuming, but also prone to human errors. The FBI scheme, on the other hand, only requires the annotator to answer at most 14 questions for each image. Furthermore, the annotator only needs to tell the relative depth ordering of two kinematically connected joints, which is intuitive and less prone to human errors, as illustrated in Fig.~\ref{fig:annotation_comparison}. Empirically, we observe that body parts with ambiguous relative depth ordering, namely, near-equal-depth joints are difficult to annotate with good accuracy. Therefore, the FBI annotation scheme only asks for ``confident" annotations from annotators. Joint pairs can be skipped and retain an ``\textit{Unknown}" label by default. 

\begin{figure}
  \centering
  \includegraphics[width= 0.90\columnwidth]{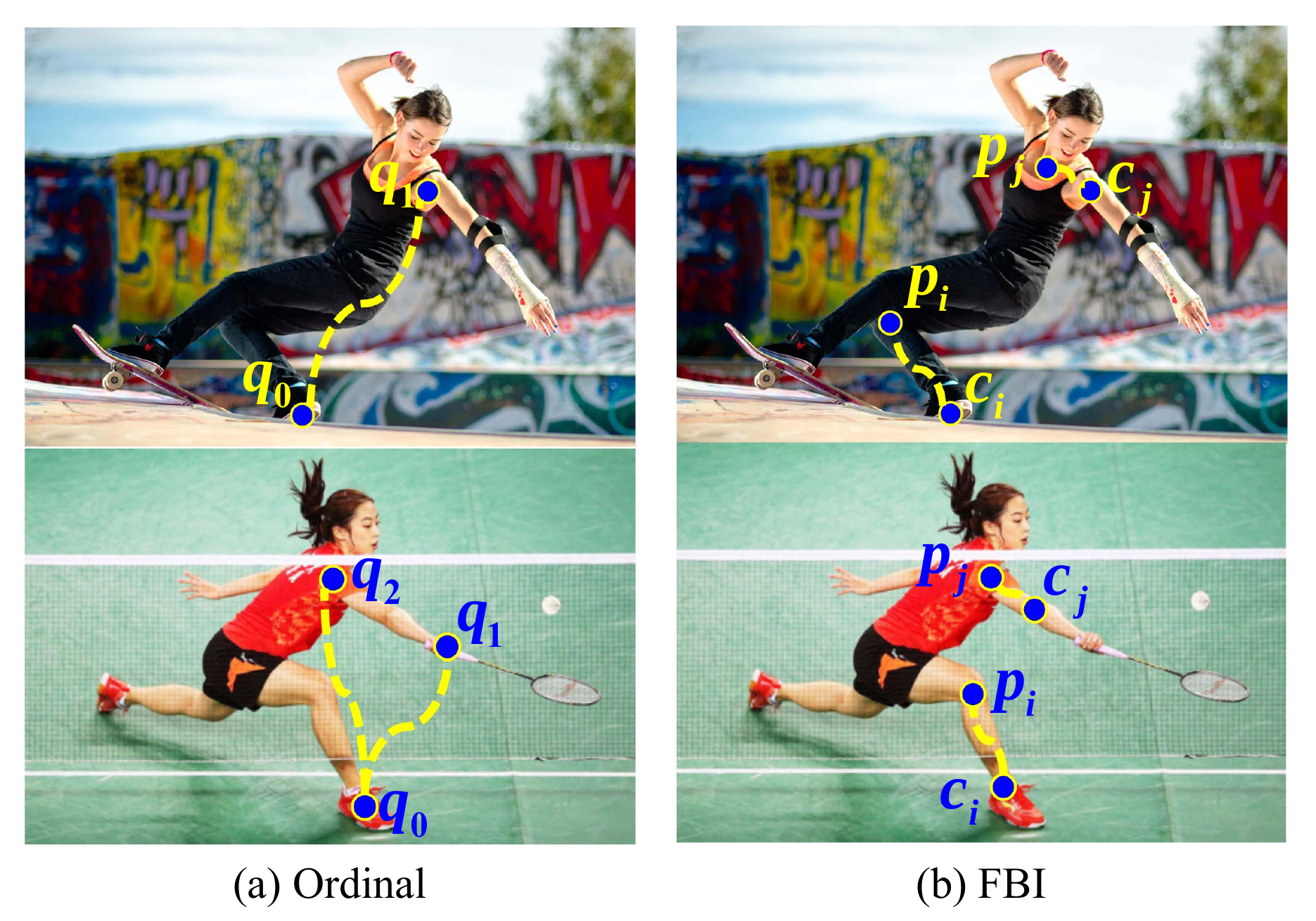}
  \caption{A simple illustration of the difference between the FBI and Ordinal annotation schemes. (a) Global relative depth ordering between disconnected joints (e.g., $(q_0,q_1)$ in the top-left image, $(q_0,q_1)$ and $(q_0,q_2)$ in the bottom-left image) need to be annotated in the Ordinal scheme, which are however challenging to annotate correctly. (b) In contrast, only local relative depth ordering between connected joints (e.g., $(p_i,c_i)$ and $(p_j,c_j)$ in the right-side images) need to be annotated in the FBI scheme.}
  \label{fig:annotation_comparison}
\end{figure}

\subsection{FBI Annotation Quality and Speed}

In order to assess the annotation quality of using the FBI scheme, 1000 images with 3D ground truth are randomly selected from the Human3.6M dataset~\cite{ionescu2014human3}. Then, they are mixed with 12K in-the-wild images for user annotations. We briefed ten first-time annotators about the FBI scheme, and collected their annotations on a total number of 13K images. For evaluation, we retrieve the annotations of all the images from the Human3.6M dataset and compare them against the ground-truth relative depth relations. We find that when the ground truth tilt angle of the skeletal bone $B_k$ with respect to the image plane is greater than $30^{\circ}$, the percentage of annotation errors is only $7.4\%$ and the percentage of skipped annotations is less than $10\%$. However, when this tilt angle is below $20^{\circ}$, both the rates of annotation errors and skipped annotations increase noticeably. This experimental study agrees with our conjecture that the body parts with small tilt angles (hence with ambiguous relative depth ordering) are much harder to annotate. 

Regarding the annotation time, on average each image takes less than 20 seconds to annotate using the FBI scheme, while the Ordinal scheme needs roughly 1 minute per image.
}


\section{Experiments} \label{sec:Experimental-Result}

In this section, we evaluate the proposed HEMlets-based human pose and shape estimation methods by conducting comprehensive experiments over the main benchmark datasets.

\begin{table*}[ht]

\setlength{\tabcolsep}{2pt}

\small

\resizebox{\textwidth}{50mm}{

\begin{tabular}{lcccccccccccccccc}
\vspace{4pt}
\\ \hline

\textbf{Protocol \#1}          & Direct        & Discuss       & Eating        & Greet         & Phone         & Photo         & Pose          & Purch.      & Sitting       & SittingD.     & Smoke         & Wait          & WalkD.        & Walk          & WalkT.        & \textbf{Avg}   \\ \hline

LinKDE~{\it et al}.~\cite{ionescu2014human3}   &132.7     &183.6    &132.3    &164.4    &162.1    &205.9    &150.6    &171.3    &151.6    &243.0    &162.1    &170.7    &177.1    &96.6    &127.9    &162.1 \\
Tome~{\it et al}.~\cite{tome2017lifting}       &65.0      &73.5     &76.8     &86.4     &86.3     &110.7    &68.9     &74.8     &110.2    &173.9    &85.0     &85.8     &86.3     &71.4    &73.1     &88.4  \\
Rogez~{\it et al}.~\cite{rogez2017lcr}		  &76.2      &80.2     &75.8     &83.3     &92.2     &105.7    &79.0     &71.7     &105.9    &127.1    &88.0     &83.7     &86.6     &64.9    &84.0     &87.7  \\
Tekin~{\it et al}.~\cite{tekin2017learning}    &54.2      &61.4     &60.2     &61.2     &79.4     &78.3     &63.1     &81.6     &70.1     &107.3    &69.3     &70.3     &74.3     &51.8    &74.3     &69.7  \\
Martinez~{\it et al}.~\cite{martinez2017simple}&53.3      &60.8     &62.9     &62.7     &86.4     &82.4     &57.8     &58.7     &81.9     &99.8     &69.1     &63.9     &67.1     &50.9    &54.8     &67.5  \\

Fang~{\it et al}.~\cite{fang2018learning}         &50.1    &54.3    &57.0    &57.1    &66.6    &73.3    &53.4    &55.7    &72.8    &88.6    &60.3    &57.7    &62.7    &47.5    &50.6    &60.4  \\
Pavlakos~{\it et al}.~\cite{pavlakos2018ordinal}  &48.5    &54.4    &54.4    &52.0    &59.4    &65.3    &49.9    &52.9    &65.8    &71.1    &56.6    &52.9    &60.9    &44.7    &47.8    &56.2  \\
S{\'a}r{\'a}ndi~{\it et al}.~\cite{sarandi2018robust}&51.2 &58.7    &51.7    &53.4    &56.8    &59.3    &50.7    &52.6    &65.5    &73.2    &56.8    &51.4    &56.6    &47.0    &42.4    &55.8  \\
Sun~{\it et al}.~\cite{sun2017integral}           &47.5    &47.7    &49.5    &50.2    &51.4    &55.8    &43.8    &46.4    &58.9    &65.7    &49.4    &47.8    &49.0    &38.9    &43.8    &49.6  \\ 
Sharma~{\it et al}.~\cite{Sharma_2019_ICCV} &48.6    &54.5   &54.2    &55.7    &62.6    &72.0    &50.5    &54.3   &70.0    &78.3    &58.1   &55.4    &61.4    &45.2    &49.7  &58.0  \\ 
Chen~{\it et al}.~\cite{chen2019weakly}      &41.1    &44.2   &44.9    &45.9   &46.5    &\textbf{39.3}   &41.6    &54.8   &73.2    &\textbf{46.2}    &48.7   &\textbf{42.1}    &\textbf{35.8}    &46.6   &38.5  &46.3 \\ 
  \hline
 Ours*~(w/o MPII 2D)   &38.5  & 45.8  & 40.3           & 54.9           & 39.5          & 45.9           &39.2            & 43.1        &49.2   &71.1           & 41.0           &53.6   &44.5  & 33.2  &34.1  &45.1\\ 
Ours                                            & \textbf{34.4}  & \textbf{42.4}  & \textbf{36.6}  & \textbf{42.1}  & \textbf{38.2}  & 39.8  & \textbf{34.7}  & \textbf{40.2}  & \textbf{45.6}  & 60.8  & \textbf{39.0}  & 42.6  &42.0  & \textbf{29.8}  & \textbf{31.7}  & \textbf{39.9}\\  

\hline

%
%

\textbf{Protocol \#2}          & Direct        & Discuss       & Eating        & Greet         & Phone         & Photo         & Pose          & Purch.      & Sitting       & SittingD.     & Smoke         & Wait          & WalkD.        & Walk          & WalkT.        & \textbf{Avg}   \\ \hline

Nie~{\it et al}.~\cite{nie2017monocular}          &90.1    &88.2    &85.7    &95.6    &103.9   &92.4    &90.4    &117.9   &136.4   &98.5    &103.0   &94.4    &86.0    &90.6    &89.5    &97.5       \\
Chen~{\it et al}.~\cite{chen20173d}               &53.3    &46.8    &58.6    &61.2    &56.0    &58.1    &41.4    &48.9    &55.6    &73.4    &60.3    &45.0    &76.1    &62.2    &51.1    &57.5       \\
Martinez~{\it et al}.~\cite{martinez2017simple}   &39.5    &43.2    &46.4    &47.0    &51.0    &56.0    &41.4    &40.6    &56.5    &69.4    &49.2    &45.0    &49.5    &38.0    &43.1    &47.7       \\
Fang~{\it et al}.~\cite{fang2018learning}         &38.2    &41.7    &43.7    &44.9    &48.5    &55.3    &40.2    &38.2    &54.5    &64.4    &47.2    &44.3    &47.3    &36.7    &41.7    &45.7  \\
Pavlakos~{\it et al}.~\cite{pavlakos2018ordinal}  &34.7    &39.8    &41.8    &38.6    &42.5    &47.5    &38.0    &36.6    &50.7    &56.8    &42.6    &39.6    &43.9    &32.1    &36.5    &41.8  \\
Yang~{\it et al}.~\cite{yang20183d}               & \textbf{26.9}    &\textbf{30.9}    &36.3    &39.9    &43.9    &47.4    &28.8    &\textbf{29.4}    &\textbf{36.9}    &58.4    &41.5    &\textbf{30.5}    &\textbf{29.5}    &42.5    &32.2    &37.7  \\ 
Sharma~{\it et al}.~\cite{Sharma_2019_ICCV} &35.3    &35.9    &45.8    &42.0    &40.9    &52.6    &36.9    &35.8    &43.5    &51.9    &44.3    &38.8   &45.5    &29.4    &34.3    &40.9  \\ \hline
Ours*~(w/o MPII 2D)   &30.8  &36.7          &31.7          &37.5           &32.5             &36.5            &29.4        &34.8    &38.5        &50.5          &33.1       &35.2   &35.9    &25.5    &27.7    &34.2 \\
Ours   &29.1  & 34.9   & \textbf{29.9}  & \textbf{32.6}  & \textbf{31.2}  & \textbf{32.3}  & \textbf{27.0}  & 33.3  & 37.6  & \textbf{45.9}  & \textbf{32.2}  &31.5  &34.5  & \textbf{22.9}  & \textbf{25.9}  & \textbf{32.1}\\ \hline

\hline

\textbf{PA MPJPE}          & Direct        & Discuss       & Eating        & Greet         & Phone         & Photo         & Pose          & Purch.      & Sitting       & SittingD.     & Smoke         & Wait          & WalkD.        & Walk          & WalkT.        & \textbf{Avg}   \\ \hline

Yasin~{\it et al}.~\cite{yasin2016dual}  &88.4    &72.5    &108.5    &110.2    &97.1    &81.6    &107.2    &119.0    &170.8    &108.2    &142.5    &86.9    &92.1    &165.7    &102.0    &108.3    \\
Sun~{\it et al}.~\cite{sun2017integral}  &36.9    &36.2    &40.6     &40.4     &41.9    &34.9    &35.7     &50.1     &59.4    &40.4    &44.9    &39.0    &\textbf{30.8}       &39.8     &36.7     &40.6    \\
Dabral~{\it et al}.~\cite{dabral2018learning} &28.0    &30.7    &39.1     &34.4     &37.1    &44.8    &28.9     &32.2     &39.3    &60.6    &39.3    &31.1    &37.8      &25.3     &28.4     &36.3    \\ \hline
Ours*~(w/o MPII 2D) &25.3  &29.1          &30.9          &30.1           &27.7             &32.7            &26.1        &28.3    &29.6        &41.9          &30.6       &26.4   &31.8    &21.7    &23.5    &29.1 \\
Ours   & \textbf{21.6}  & \textbf{27.0}  & \textbf{29.7}  & \textbf{28.3}  & \textbf{27.3}  & \textbf{32.1}  & \textbf{23.5}  & \textbf{30.3}  & \textbf{30.0}  & \textbf{37.7}  & \textbf{30.1}  & \textbf{25.3}  & 34.2 &\textbf{19.2}  & \textbf{23.2}  & \textbf{27.9}\\ \hline
\end{tabular}

}

\vspace{2pt}
\caption{Quantitative comparisons of the mean per-joint position error (MPJPE) on Human3.6M~\cite{ionescu2014human3} under Protocol~\#1 and Protocol~\#2, as well as using PA MPJPE as the evaluation metric. Similar to most of the competing methods (e.g.,~\cite{sun2017integral,pavlakos2018ordinal,yang20183d,dabral2018learning,tekin2017learning,fang2018learning}), our models were trained on the Human3.6M dataset and used also the extra MPII 2D pose dataset~\cite{Andriluka}. We also report the results without using the extra MPII 2D pose dataset.} 

\label{table:protocols}
\end{table*}

\subsection{3D Human Pose Estimation}
We perform quantitative evaluation on three benchmark datasets: Human3.6M~\cite{ionescu2014human3}, HumanEva-I~\cite{sigal2010humaneva} and MPI-INF-3DHP~\cite{mehta2017monocular}. 
Ablation study is conducted to evaluate our design choices. 
We demonstrate that the proposed method shows superior generalization ability to 
in-the-wild images. 

\subsubsection{Datasets and Evaluation Protocols}

{\bf Human3.6M.}
Human3.6M~\cite{ionescu2014human3} contains 3.6 million RGB images captured by a MoCap System in 
an indoor environment, in which 7 professional actors were performing 
15 activities such as walking, eating, sitting, making a phone call and 
engaging in a discussion, etc. 
We follow the standard protocol as in~\cite{martinez2017simple, pavlakos2017coarse}, 
and use 5 subjects (S1, S5, S6, S7, S8) for training and the rest 2 subjects (S9, S11) 
for evaluation (referred to as Protocol~\#1). 
Some previous works reported their results with 6 subjects 
(S1, S5, S6, S7, S8, S9) used for training and only S11 for evaluation~\cite{yasin2016dual,sun2017integral,dabral2018learning} (referred to as Protocol~\#2). Despite {\it not} using S9 also as training data, 
we compare our results with these methods.

{\bf HumanEva-I.} 
HumanEva-I~\cite{sigal2010humaneva} is one of the early datasets for evaluating 3D human poses. 
It contains fewer subjects and actions compared to Human3.6M. 
Following~\cite{Bo2010Twin}, we train a single model on 
the training sequences of Subject 1, 2 and 3, and evaluate on 
the validation sequences.

{\bf MPI-INF-3DHP.} 
This is a recent 3D human pose dataset which includes both indoor and outdoor scenes~\cite{mehta2017monocular}. 
Without using its training set, we evaluate our model trained from Human3.6M only 
on the test set. The results are reported using the 3DPCK and the AUC metric~\cite{Andriluka,mehta2017monocular,pavlakos2018ordinal}.

{\bf Evaluation metric.}
We follow the standard steps to align the 3D pose prediction with the groundtruth by 
aligning the position of the central hip joint, and use 
the \textit{Mean Per-Joint Position Error}~(MPJPE) between 
the groundtruth and the prediction as evaluation metrics. 
In some prior works~\cite{yasin2016dual,sun2017integral,dabral2018learning}, 
the pose prediction was further aligned with the 
groundtruth via a rigid transformation. The resulting MPJPE is termed as \textit{Procrustes Aligned}~(PA) MPJPE. 

\subsubsection{Results and Comparisons}
\label{sec:stoa}
{\bf Human3.6M.} 
We compare our method against  state-of-the-art under three protocols, and the quantitative 
results are reported in Table~\ref{table:protocols}. As can be seen,
our method outperforms all competing methods on nearly all action subjects 
for the protocols used. It is worth mentioning that our method 
makes considerable improvements on some challenging actions for 3D pose 
estimation such as {\it Sitting} and {\it Walking}. 
Thanks to HEMlets learning, our method demonstrates a clear advantage for handling 
complicated poses. 

With a simple network architecture and little parameter 
tuning, we produce the most competitive results compared to previous 
works with carefully designed networks powered by e.g., adversarial training schemes 
or prior knowledge. On average, we improve the 3D pose prediction accuracy 
by $20\%$ than that reported in Sun~{\it et al}.~\cite{sun2017integral} under Protocol \#1. 
We also report our performance using PA MPJPE as the evaluation metric, and 
compare with these methods that make use of S9 as additional training data. We 
still outperform all of them across all action subjects, even {\it without} 
utilizing S9 for training.  {For fair comparison with the competing methods in Table~\ref{table:protocols}, our models were trained similarly on the Human3.6M dataset and used also the extra MPII 2D pose dataset. We have also trained our method by additionally using the FBI dataset, and the 3D pose prediction accuracy obtained further improves as the MPJPE goes from 39.9mm down to 36.5mm.} 

\begin{table}[t]
\small
\setlength{\tabcolsep}{2pt}
  \centering
	
  \begin{tabular}{l|ccc|ccc|c}
\hline

\multirow{2}{*}{Approach}&
\multicolumn{3}{c|}{Walking}&
\multicolumn{3}{c|}{Jogging}  & \multirow{2}{*}{Avg}\\ \cline{2-7}
									&S1    &S2    &S3    &S1    &S2    &S3     \\ \hline

Simo-Serra~{\it et al}.~\cite{simo2013joint}         &65.1  &48.6  &73.5  &74.2  &46.6  &32.2  &56.7 \\ 
Moreno-Noguer~{\it et al}.~\cite{moreno20173d}       &19.7  &13.0  &24.9  &39.7  &20.0  &21.0  &26.9 \\ 
Martinez~{\it et al}.~\cite{martinez2017simple}      &19.7  &17.4  &46.8  &26.9  &18.2  &18.6  &24.6 \\ 
Fang~{\it et al}.~\cite{fang2018learning}            &19.4  &16.8  &37.4  &30.4  &17.6  &16.3  &22.9 \\
Pavlakos~{\it et al}.~\cite{pavlakos2018ordinal}     &18.8  &12.7  &29.2  &\textbf{23.5}  &15.4  &14.5  &18.3 \\ \hline
Ours                                           &\textbf{13.5}  &\textbf{9.9}  &\textbf{17.1}   &24.5 &\textbf{14.8}  &\textbf{14.4}  &\textbf{15.2} \\ \hline

\end{tabular}
\vspace{4pt}
\caption{Detailed results on the validation set of HumanEva-I~\cite{mehta2017monocular}. }
\label{table:HumanEva-I}
\end{table}

{\bf HumanEva-I.}
With the same network architecture where {\it only} the HumanEva-I dataset is used for training, our results are reported in Table~\ref{table:HumanEva-I} under the popular protocol~\cite{simo2013joint,moreno20173d,martinez2017simple,fang2018learning,pavlakos2018ordinal}. Different from these approaches~\cite{pavlakos2018ordinal,moreno20173d,martinez2017simple,fang2018learning} which used extra 2D datasets~(e.g., MPII) or pre-trained 2D detectors~(e.g., CPM~\cite{wei2016convolutional}), our method still outperforms previous approaches.

\begin{table}[t]
\small
\setlength{\tabcolsep}{2pt}
  \centering
	
  \begin{tabular}{lccccc}
\hline

\multirow{3}{*}{Approach}

&Studio  & Studio &

\multirow{2}{*}{Outdoor  }&
\multirow{2}{*}{All  }&
\multirow{2}{*}{All  } \\

&GS &no GS \\ 

\cline{2-6}

&3DPCK   &3DPCK   &3DPCK   &3DPCK  &AUC\\

\hline

Mehta~{\it et al}.~\cite{mehta2017monocular}     &70.8    &62.3    &58.8    &64.7    &31.7\\
Zhou~{\it et al}.~\cite{zhou2017towards}         &71.1    &64.7    &72.7    &69.2    &32.5\\
Pavlakos~{\it et al}.~\cite{pavlakos2018ordinal} &\textbf{76.5}    &63.1    &77.5    &71.9    &35.3\\
\hline
Ours                                       &75.6    &\textbf{71.3}    &\textbf{80.3}    &\textbf{75.3}    &\textbf{38.0}\\
\hline
\end{tabular}
\vspace{4pt}
\caption{Detailed results on the test set of MPI-INF-3DHP~\cite{mehta2017monocular}. No training data from this dataset was used to train our model.}
\label{table:3dhpResults}
\end{table}

{\bf MPI-INF-3DHP.}
We evaluate our method on the MPI-INF-3DHP dataset using two metrics, the PCK and AUC. The results are generated by the model we trained for Human3.6M. In Table~\ref{table:3dhpResults}, we compare with three recent methods which are not trained on this dataset. Our result of ``Studio GS" is one percentage lower than~\cite{pavlakos2018ordinal}. But our method outperforms all these methods with particularly large margins for the ``Outdoor" and ``Studio no GS" sequences.

\subsubsection{Ablation Study}
\label{sec:ablation}

We study the influence on the final estimation performance of different choices made in our network design and training procedure.

{\bf Alternative intermediate supervision.}
First, We examine the effectiveness of using HEMlets supervision. 
We evaluate the model trained without any intermediate supervision (Baseline), 
with 2D heatmap supervision only, with HEMlets supervision only, and 
with both 2D heatmap supervision and HEMlets supervision (Full). 
All of these design variants are evaluated with the same experimental 
setting (including training data, network architecture and $\mathcal{L}^{\rm 3D}_{\lambda}$ loss 
definition) under Protocol \#1 on Human3.6M. 


\begin{table}[t]
\small
\renewcommand\tabcolsep{3.0pt} 
\begin{center}
\begin{tabular}{llcc}
\hline
Method & Supervision & H3.6M \#1  & H3.6M \#1$^*$ \\

\hline

 Baseline               &  $\mathcal{L}^{\rm 3D}_{\lambda}$& 47.1  &55.3\\
w/ 2D heatmaps                    & $\mathcal{L}^{\rm 3D}_{\lambda} + \mathcal{L}^{\rm 2D}$ & 44.2  & 49.9\\
w/   HEMlets                & $\mathcal{L}^{\rm 3D}_{\lambda} + \mathcal{L}^{\rm HEM}$& 42.6		& 46.0\\
   Full                & $\mathcal{L}^{\rm 3D}_{\lambda} + \mathcal{L}^{\rm HEM} + \mathcal{L}^{\rm 2D}$ & 39.9 & 45.1\\

\hline
\end{tabular}
\end{center}
\caption{Ablative study on the effects of alternative intermediate supervision evaluated on Human3.6M using Protocol~\#1. The last column~$^*$ reports the results using only the Human3.6M dataset for training (without using the extra MPII 2D pose dataset).}
\label{table:ablationStudy}
\end{table}


The detailed results are presented in Table~\ref{table:ablationStudy}. Using 2D heatmaps supervision for training, the prediction error is reduced by 3.0mm compared to the baseline. The HEMlets supervision provided 1.7mm lower mean error compared to the 2D heatmaps supervision. This validates the effectiveness of the intermediate supervision. By combining all these choices, our approach using HEMlets with 2D heatmap supervision achieves the lowest error. Without using the extra MPII 2D pose dataset, we repeated this study. Similar conclusions can still be drawn. But the gap between w/ HEMlets (excluding $\mathcal{L}^{\rm 2D}$, 46.0mm) and Full (45.1mm) shrinks, suggesting the strength of the HEMlets representation in encoding both 2D and (local) 3D information.

\begin{figure}[t]
\centering
\includegraphics[width = \columnwidth ]{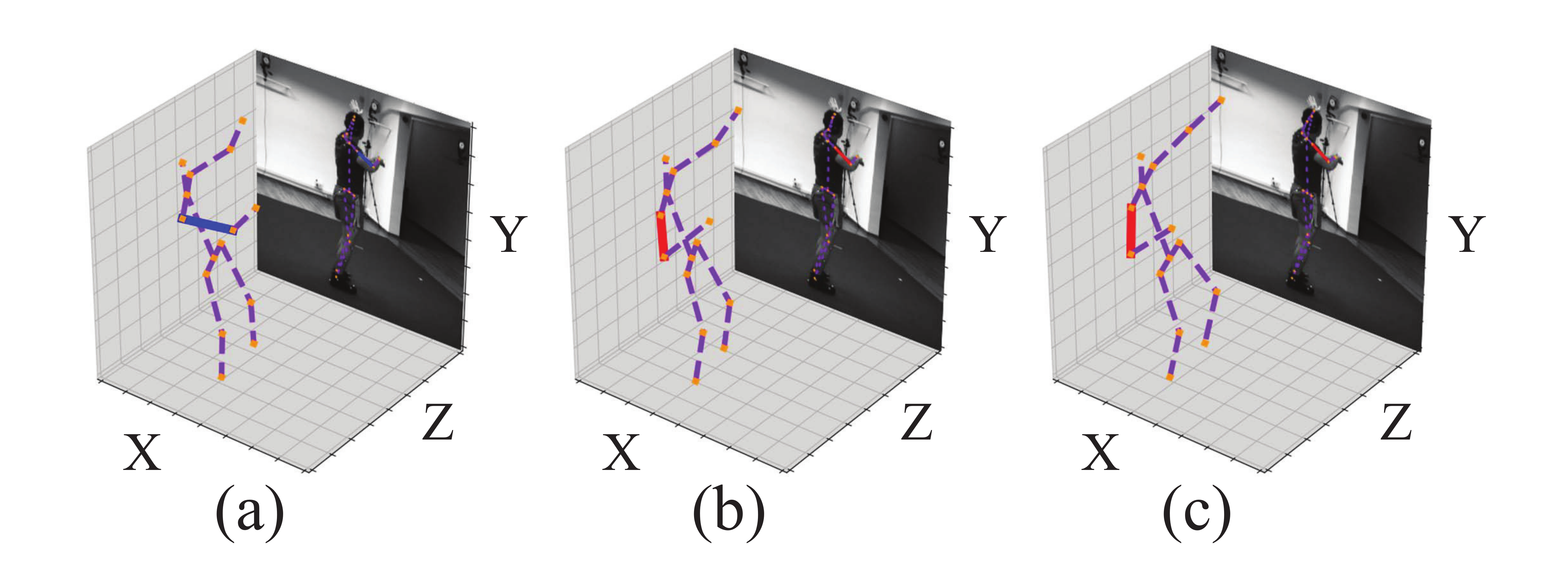}
\caption{An example image with the detected joints overlaid and shown from a novel view, using different methods: (a) $\mathcal{L}^{\rm 3D}_{\lambda} + \mathcal{L}^{\rm 2D}$ (2D error:~15.2; 3D joint error:~{\bf81.3mm}). (b) $\mathcal{L}^{\rm 3D}_{\lambda} + \mathcal{L}^{\rm 2D} + \mathcal{L}^{\rm HEM}$ (2D error:~13.0; 3D error:~{\bf41.2mm}). (c) Ground-truth. HEMlets learning helps fixing local part errors, see the blue skeletal part in~(a) versus the red skeletal part in~(b).}
\label{fig:example}
\end{figure}

To further illustrate the effectiveness of HEMlets representation, we provide a visual comparison in Fig.~\ref{fig:example}. Though the 2D joint errors of the two estimations are quite close, the method with HEMlets learning significantly improves the 3D joint estimation result and fixes the gross limb errors.

Regarding the runtime, tested on a NVIDIA GTX 1080 GPU, our full model (with a total parameter number of 47.7M) takes 13.3ms for a single forward inference, while the baseline model (with 34.3M parameters) takes 8.5ms.

\begin{figure}[h]
  \centering
  \includegraphics[width=0.80\columnwidth]{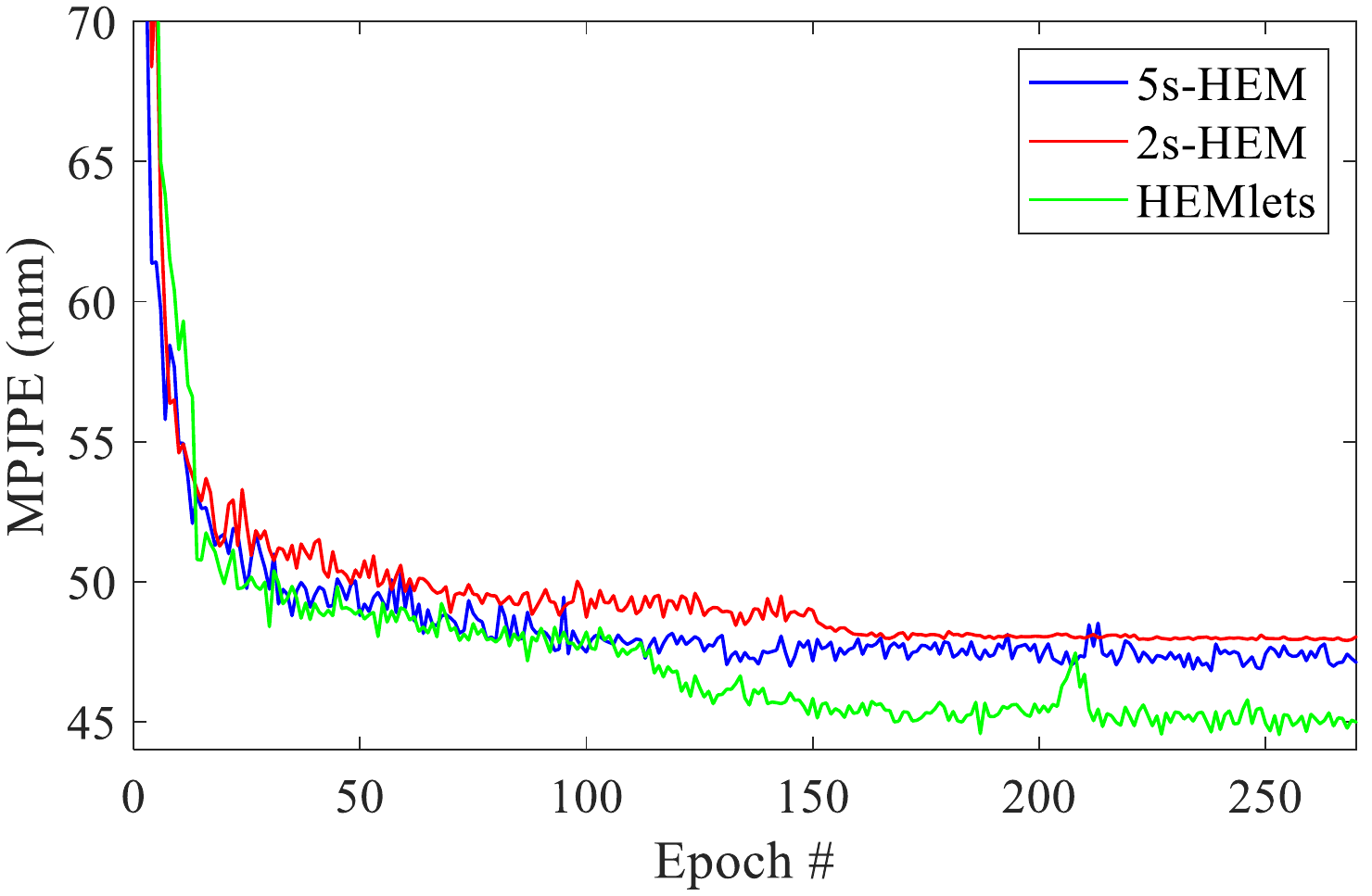} 
\caption{The MPJPE of the validation set of 5s-HEM, 2s-HEM and HEMlets, respectively. All are trained with the Human3.6M dataset. } 
\label{fig:Fiv-tri-state result}
\end{figure}

{\bf Variants of HEMlets.}
We next experimented with some variants of HEMlets on Human3.6M and MPII 2D pose datasets. In the first variant, we use five-state heatmaps, 
referred to as \textit{5s-HEM}, where the child joint is placed to different layers of the 
heatmaps according to the angle of the associated skeletal part with respect to the imaging plane. 
Specifically, we 
define the five states corresponding to the $(-90^\circ,-60^\circ)$, $(-60^\circ, -30^\circ)$, 
$(-30^\circ, 30^\circ)$, $(30^\circ, 60^\circ)$ and $(60^\circ, 90^\circ)$ range, respectively. 
In the second variant, we place a pair of joints in the negative and positive polarity
heatmaps respectively according to their depth ordering (i.e., the closer/farther joint will appear in the positive/negative polarity heatmap). If their depths are roughly the same, they are co-located in the zero polarity heatmap. We refer to this variant as \textit{2s-HEM}.  We trained 5s-HEM, 2s-HEM and HEMlets with the Human3.6M dataset only. A comparison on the MPJPE of the validation set is given in Fig.~\ref{fig:Fiv-tri-state result}.

The other two variants produce inferior convergence compared to HEMlets under the same experiment setting. 

\begin{table}[h]
\begin{center}
\begin{tabular}{lcc}
\hline
Dataset     &3DPCK\\
\hline
Base                     & 75.3    \\ 
w/  Ordinal~\cite{pavlakos2018ordinal}             & 76.1    \\
w/  FBI~\cite{shi2018fbi}       	    & \textbf{76.9}	   \\

w/  FBI~\cite{shi2018fbi} + Ordinal~\cite{pavlakos2018ordinal}     & 76.5    \\
\hline
\end{tabular}
\end{center}
\caption{Evaluation of 3DPCK scores by adding different augmenting datasets that provide relative depth ordering annotations. Base denotes using the base datasets~(Human3.6M and MPII). }
\label{table:DataSetEffectiveness}
\end{table}

\begin{figure}[h]
  \centering
  \includegraphics[width=0.95\columnwidth]{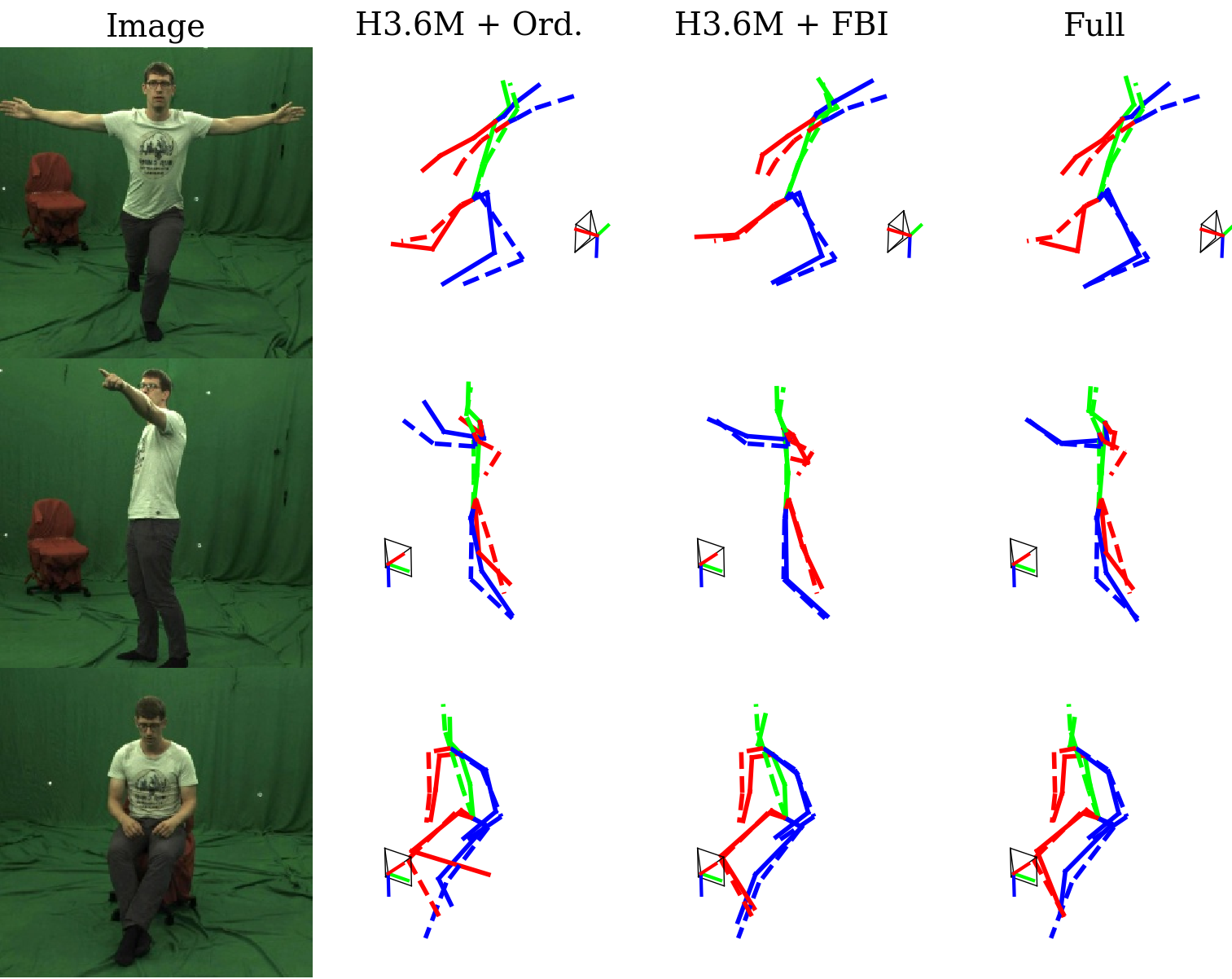} 
  \caption{The qualitative results for some examples of MPI-INF-3DHP~\cite{mehta2017monocular}, using different additional datasets. For each example, we present the input RGB image, the 3D human pose predicted by three different models. The groundtruth pose is shown in dashed line.} 
\label{fig:DifDataset result}
\end{figure}

{\bf Augmenting datasets.}
Many state-of-the-art approaches use a mixed training strategy for 3D human pose estimation. 
In addition to exploiting Human3.6M and MPII datasets, we study the effect of using 
augmenting datasets such as Ordinal~\cite{pavlakos2018ordinal} and FBI~\cite{shi2018fbi} 
for training. Firstly, we adapt the annotations of Ordinal and FBI datasets to the required form of HEMlets. 
Then we train our model using different combinations of these additional datasets. 
The comparisons on the MPI-INF-3DHP dataset~\cite{mehta2017monocular} are reported in Table~\ref{table:DataSetEffectiveness}. We find augmenting datasets 
slightly increase the 3DPCK score for the trained model. Interestingly, training with FBI annotations 
attains a better 3DPCK score than Ordinal annotations. We suspect this is 
due to the amount of manual annotation errors related to different annotation schemes. In Fig.~\ref{fig:DifDataset result}, we also provide some visual examples to compare the effectiveness of different augmenting datasets. One can find that the model fine-tuned with the FBI dataset produces better predictions than the ones trained additionally with Ordinal~\cite{pavlakos2018ordinal}.


\begin{figure*}[ht]
  \centering
  \includegraphics[width=0.95\linewidth]{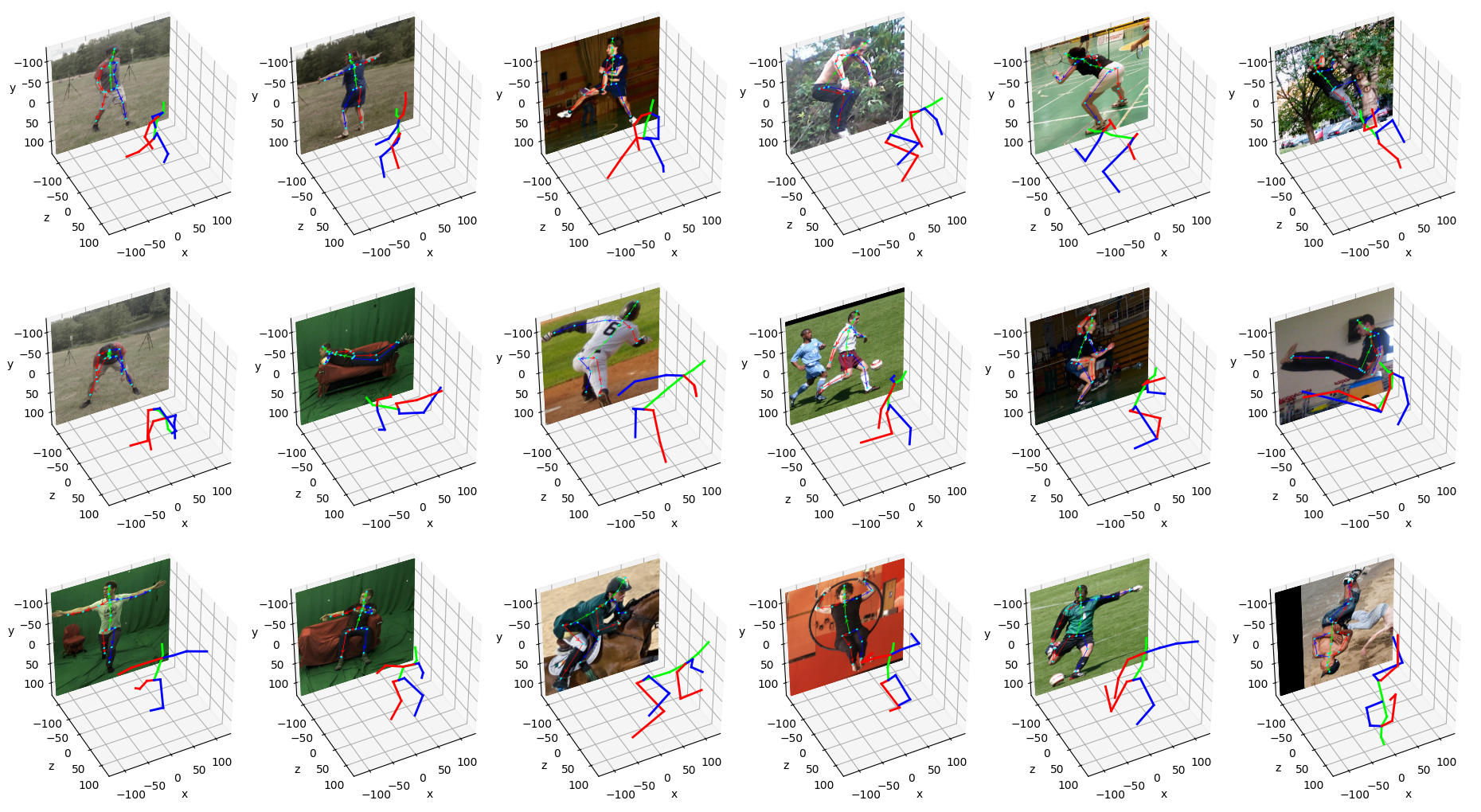}
  \caption{Qualitative results on different validation datasets: the first two columns are from the test dataset of 3DHP~\cite{mehta2017monocular}. The other columns are from Leeds Sports Pose~(LSP)~\cite{johnson2010clustered}. Our approach produces visually correct results even on challenging poses~(last column). }
\label{fig:wild}
\end{figure*}

{\bf Generalization.}
For an evaluation of in-the-wild images from Leeds Sports Pose~(LSP)~\cite{johnson2010clustered} and the validation set of MPI-INF-3DHP~\cite{mehta2017monocular}, we list some visual results predicted by our approach. As shown in Fig.~\ref{fig:wild}, even for challenging data~(e.g., self-occlusion, upside-down), our method yields visually correct pose estimations for these images.  

\subsection{3D Human Body Model Recovery}
In this part, we evaluate the proposed human body recovery method of regressing the SMPL parameters on three public datasets i.e., SURREAL~\cite{varol2017learning}, UP-3D~\cite{lassner2017unite} and 3DPW\cite{von2018recovering}. Before the experimental studies, we first give an introduction to the datasets and related evaluation protocols. 

\subsubsection{Datasets and Evaluation Protocols}
\label{sec:dataset}
{\bf SURREAL.} SURREAL~\cite{varol2017learning} contains 6M frames from 1,964 video sequences of 115 subjects, where the images are photo-realistic renderings of people under large variations in shape, texture, viewpoint and pose. Because these synthetic bodies are created using SMPL body models, the corresponding model parameters are used as groundtruth for training a human body regression model. 

{\bf UP-3D.}
To construct the UP-3D dataset~\cite{lassner2017unite}, the authors have collected a large number of real images mostly from the 2D human pose datasets~(i.e., MPII~\cite{Andriluka} and LSP~\cite{johnson2010clustered}). There are two main steps to produce the final dataset. First, a SMPLify optimization~\cite{bogo2016keep} is applied to obtain the 3D human body mesh results. Then, those inaccurate body fitting results were inspected and discarded manually by humans. As a result, 8,515 images with the fitted SMPL body parameters are obtained, among which 7,818 images are used for training and 1,389 for testing.


{\bf 3DPW.} Recently, the work of~\cite{von2018recovering} presented a new dataset which is captured under in-the-wild environment. Specifically, a moving hand-held camera is used for recording RGB frames while IMUs are attached on actors to capture poses. In total, 60 video sequences (more than 51,000 frames) of 5 subjects are captured, where 7 actors with 18 different clothing styles are asked to perform different activities, such as  walking, playing golf and etc.


{\bf Evaluation metric.} We follow the standard protocols, as detailed in~\cite{yao2019densebody} to conduct evaluations. When dealing with the datasets of SURREAL and UP-3D, to measure the accuracy of the inferred body mesh, the average per-vertex Euclidean distance between it and the groundtruth is used (which is referred to as ``surface''). We also report the accuracy of the output 3D pose, where the average per-joint Euclidean distance between the estimated pose (with the hip joint aligned) and the groundtruth is used (which is referred to as ``joint''). For the dataset of 3DPW, we follow the works of~\cite{kanazawa2018end,kolotouros2019spin} to evaluate the reconstruction error of 3D poses, which is noted as ``Rec. Error''.
Basically, ``Rec. Error'' is computed as follows in SPIN~\cite{kolotouros2019spin}: it measures the average per-joint Euclidean distance between the predicted 3D human pose and the groundtruth after a global alignment post-process, where the predicted pose is regressed from the inferred 3D body mesh. In addition, following the work of~\cite{xu2019denserac}, the recovered 3D meshes are also projected onto a 2D image plane for evaluating the accuracy of the mask and part segmentation. By doing so, mIoU and F1 scores are reported.


\begin{table}[t]
\footnotesize
\setlength{\tabcolsep}{2pt}
  \centering
	
  \begin{tabular}{l|cc|cc|cc|c}
\hline

\multirow{2}{*}{Approach}& 
\multicolumn{2}{|c|}{Human3.6M}&
\multicolumn{2}{|c|}{SURREAL}&
\multicolumn{2}{|c|}{UP-3D}&
3DPW\\  \cline{2-8}

					&Pro.\#1  &Pro. \#2				&surface    &joint    &surface    &joint   &Rec. Error    \\ \hline

Pavlakos~{\it et al}.~\cite{pavlakos2018learning}   & -   &75.9  & -   &-  &117.7  &-  &- \\ 
HMR~\cite{kanazawa2018end}                    & 88.0   &59.1  & -   &-  &-  &-  &81.3  \\ 

BodyNet ~\cite{varol2018bodynet}                                                    & -   &-         &73.6   &-  &- &-  &-  \\ 
SMPLR~\cite{madadi2018smplr}                & 56.5  & 46.3      &74.5   &46.1  &- &-  &-  \\ 
DenseRaC~\cite{xu2019denserac} & 76.8   &-            &-   &- &- &-  &-   \\  
TexturePose~\cite{pavlakos2019texturepose}& 51.3   & 49.7           &-   &- &- &- &-  \\  

DenseBody~\cite{yao2019densebody}                                              & 47.3   &38.1             &54.2   &40.1 &91.7 &71.4  &-  \\   

SPIN (SPIN*)~\cite{kolotouros2019spin }&-    & 41.1          &-   &- &- &- 
&66.3 (59.2*)  \\  \hline

\strut Ours                                & \textbf{39.9}  &\textbf{32.1}           &\textbf{53.3}  &\textbf{37.7} 
&\textbf{79.8}   &\textbf{67.5}  &\textbf{58.8} \\ \hline

\end{tabular}
\vspace{5pt}
\caption{Quantitative comparisons of fully body model recovery results over different datasets. * denotes the version that also applies the SMPLify optimization~\cite{bogo2016keep} as post-processing. All the numbers listed are in mm.
}

\label{table:Shape}
\end{table}


\begin{table}
\small
\setlength{\tabcolsep}{3pt}
  \centering

  \begin{tabular}{l|c|c|c|c } 
\hline
\multirow{2}{*}{Approach}&
\multicolumn{2}{|c|}{FB Seg.}&
\multicolumn{2}{|c}{Part Seg.}  \\  \cline{2-5}

					&Accuracy   &F1				&Accuracy   &F1       \\ \hline

SMPLify oracle~\cite{lassner2017unite}   & 92.17           &0.88       &88.82     &0.67       \\ 
SMPLify~\cite{bogo2016keep}                   & 91.89           &0.88       &87.71     &0.64      \\ 
HMR~\cite{kanazawa2018end}              & 91.67          &0.87        &87.12      &0.60        \\ 

SPIN~\cite{kolotouros2019spin }           &91.07          &0.86          &88.48   &0.65  \\ 
SPIN*~\cite{kolotouros2019spin}        &91.83           &0.87         &89.41    &0.68  \\

BodyNet ~\cite{varol2018bodynet}                  & \textbf{92.80}  &0.84        &-   &-   \\ 
DenseRaC~\cite{xu2019denserac}                   & 92.40  &0.88        &87.90   &0.64   \\ 

\hline
Ours                                 & 92.30     &0.88           &\textbf{90.18}  &\textbf{0.71}    \\ 
Ours*                               &\textbf{93.67}          &\textbf{0.90}  &\textbf{91.19} &\textbf{0.74}   \\ \hline

\end{tabular}
\vspace{4pt}
\caption{Quantitative comparisons between our method and existing ones on foreground and part segmentation of the recovered full body mesh on the UP-3D dataset. * denotes the version that also applies the SMPLify optimization~\cite{bogo2016keep} as post-processing.}
\label{table:SegEva}
\end{table}

\subsubsection{Results and Comparisons}
Next, we report the evaluation results and also compare them with state-of-the-art methods both quantitatively and qualitatively. 

{\bf Quantitative comparisons.} In Table~\ref{table:Shape}, we numerically compare our method to existing leading approaches on the evaluation metrics presented in Sect.~\ref{sec:dataset}. As can be seen, our method produces the best accuracy for both the output skeleton joints and the generated body mesh. Table~\ref{table:SegEva} also lists the accuracy of the foreground and the part segmentation, given the generated body mesh. Our proposed method again gives the best performance. It is noteworthy that the part segmentation F1 score of our method evaluated on the UP-3D dataset exceeds 0.70 for the first time.

{\bf Qualitative comparisons.} We also conduct qualitative comparisons between our method and some of existing methods, as shown in Fig.~\ref{fig:shapecomp}. Here, HMR~\cite{kanazawa2018end} and SPIN~\cite{kolotouros2019spin} are selected as two representative body mesh recovery approaches. Given an input image, the output body mesh of each method is shown in two views. It can be observed that our method performs better than HMR and SPIN, even when the human pose is challenging. 


{\color{blue}


\begin{figure*}[t]
\centering
\includegraphics[width=16.5cm]{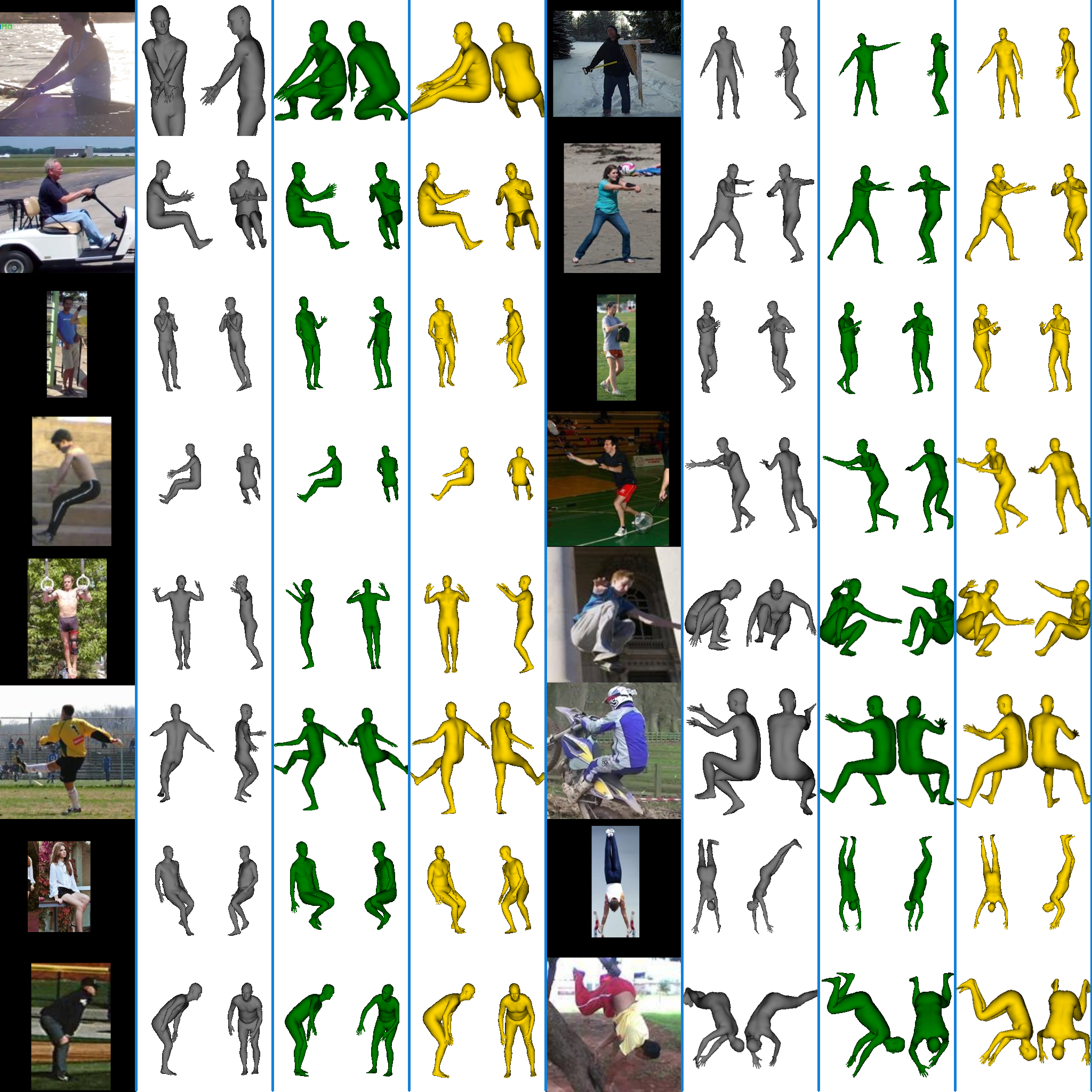}
\caption{Qualitative comparisons of our method with some existing ones on human body model recovery. For each example, the input image is first shown, which is followed by the results of HMR~\cite{kanazawa2018end},  SPIN~\cite{kolotouros2019spin} and ours. For each resulting body mesh, two views are provided for visualization.
}
\label{fig:shapecomp}
\end{figure*}
}

\begin{figure}[t]
\centering
\includegraphics[width=\linewidth]{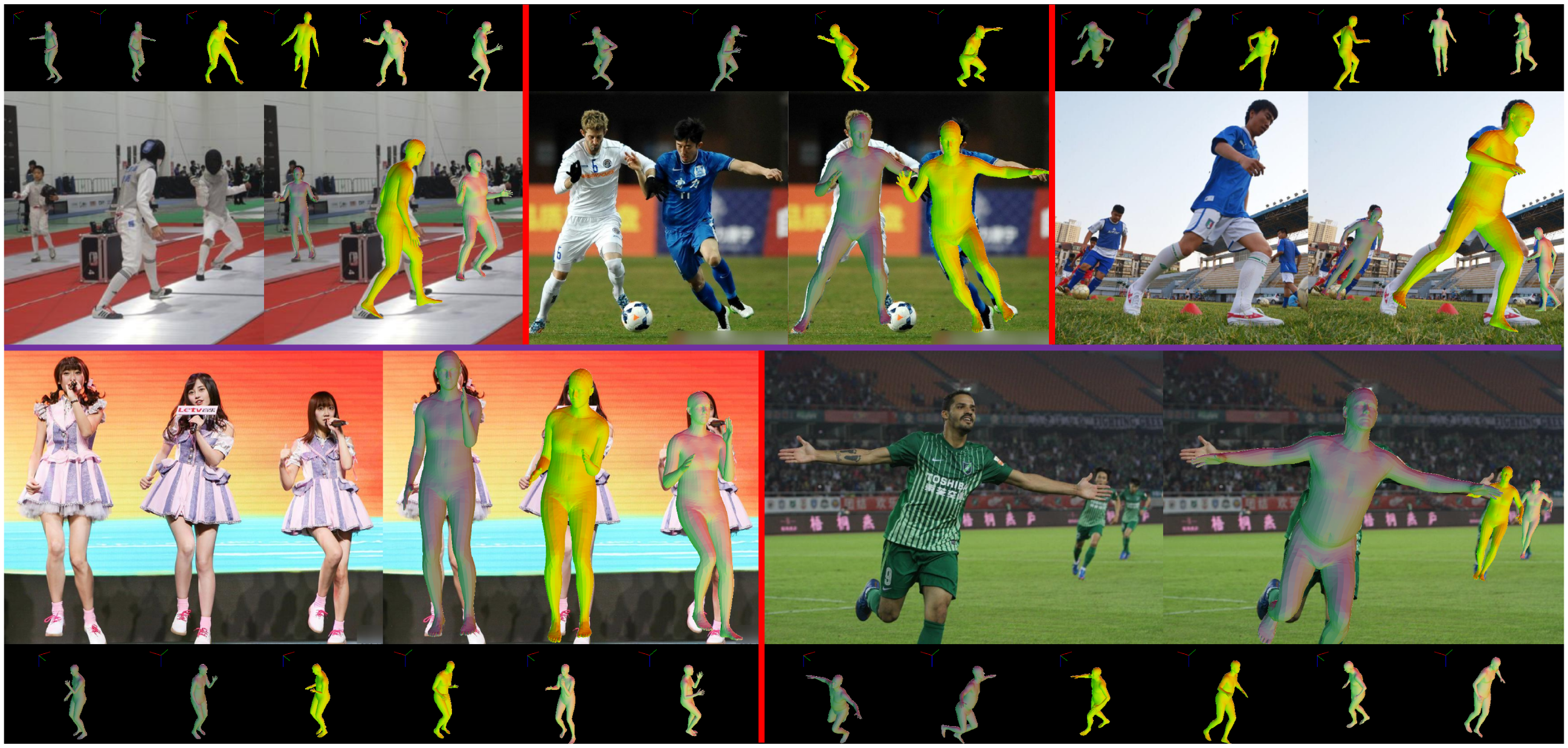}
\caption{The results of the proposed approach on multi-person scenarios.}
\label{fig:multi-shape}
\end{figure}

{
{\bf Ablation study.}
For the ablative analysis of 3D human body shape regression, we trained two alternative models to evaluate the necessity of each branch of the proposed HEMlets-based body regression method depicted in Fig.~\ref{fig:shapeModule}: 1) by only using the 3D joints branch in Fig. 5, and 2) by only using the high-level feature extraction branch in Fig.~\ref{fig:shapeModule}. The results are reported in Table~\ref{table:shape}. It can be observed that the 3D joints regression component plays a more important role for the task of 3D body shape regression.

\begin{table}[t]
\begin{center}
\begin{tabular}{lc}
\hline
       Method         &Rec. Error measured on 3DPW\\
\hline
w/ only high-level features                                     &{63.1}   \\
w/ only the 3D joints                                                &{60.3}   \\ 
Full                                                            &\bf{58.8}    \\
\hline

\end{tabular}
\end{center}
\caption{Evaluating the impact of each branch in Fig.~\ref{fig:shapeModule} on human body estimation. }
\label{table:shape}
\end{table}
}

\subsubsection{Extended Studies}

\begin{table}[t]
\begin{center}
\begin{tabular}{lc}
\hline
       Method         &Rec. Error measured on 3DPW\\
\hline
Proposed model      &58.8    \\
w/  groundtruth shape $\beta^{\rm gt}$
                    &57.2    \\ 
w/  groundtruth pose $\theta^{\rm gt}$            &\bf{9.4}   \\

\hline
\end{tabular}
\end{center}
\caption{Evaluation of the impact of learning $\theta$ and $\beta$ on human body estimation. }
\label{table:Shape_abalation}
\end{table}

\begin{figure}[t]
\center
\includegraphics[width=0.96\linewidth]{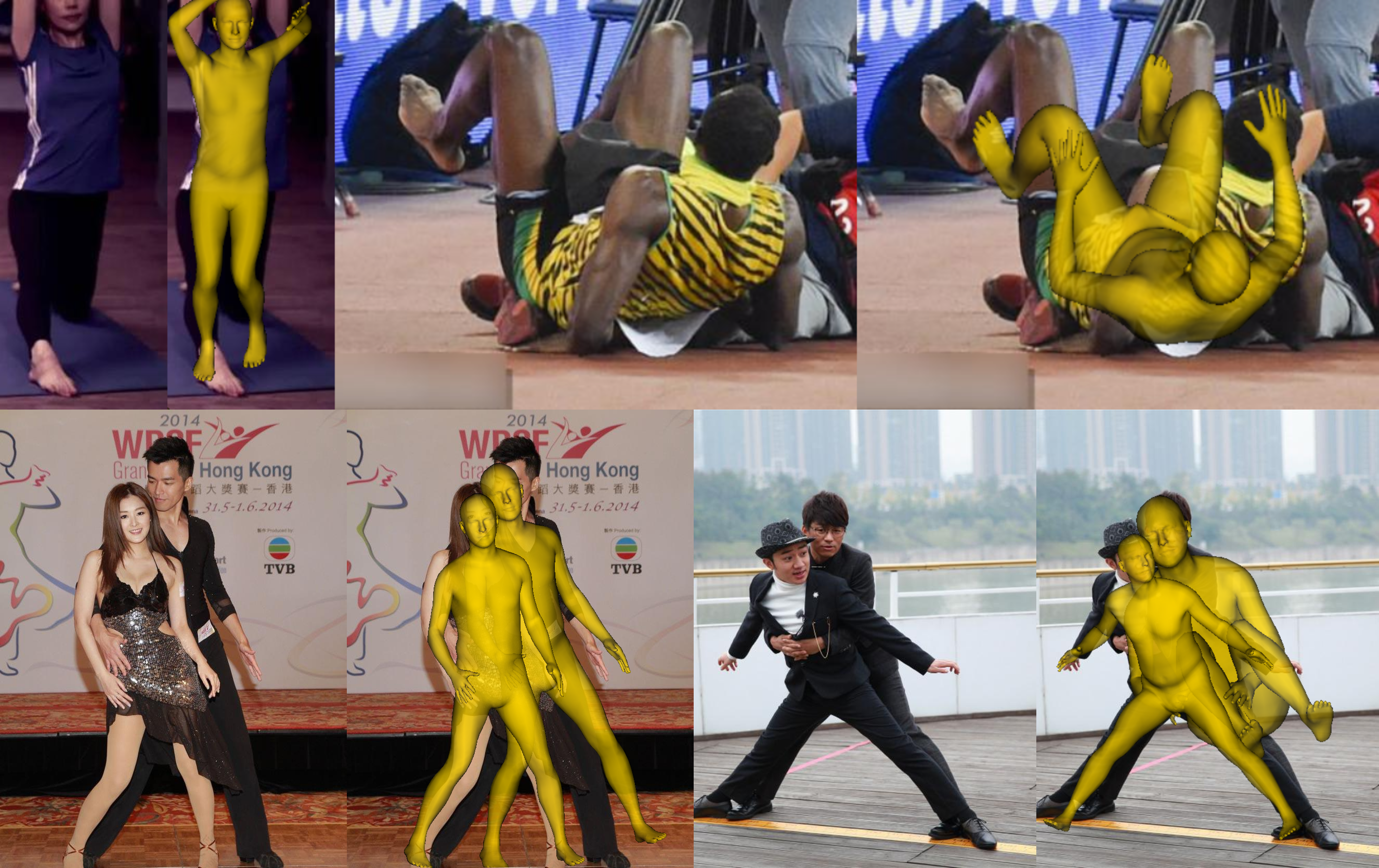}
\caption{Some failure cases of the proposed approach.}
\label{fig:failurecase}
\end{figure}

We make a few extended studies to further understand the proposed HEMlets-based human pose and shape estimation method.

\indent{\bf How does pose accuracy affect body recovery?} As the accuracy of the output body mesh relies on both the estimated pose and the shape parameters. An interesting question is which factor affects more. To reach an answer, we run two alternative versions of our full model on the 3DPW dataset: 1) replacing its estimated shape parameter with the groundtruth shape, and 2) replacing the estimated pose parameter with the groundtruth pose. The results are reported in Table~\ref{table:Shape_abalation}. As one can see, the accuracy of pose estimation has a greater impact. This suggests 3D pose estimation is critical and provides more significant contributions to the task of human body mesh recovery from a single color image. 


\indent{\bf Multi-person 3D pose and shape.} Fig.~\ref{fig:multi-shape} shows our method can also work well for the multi-person scenarios. To do that, we firstly employ the code of OpenPose~\cite{cao2016realtime} to detect person instances. Each instance is then cropped, to which the proposed HEMlets PoSh approach is applied for individual 3D body model inference.     


{\bf Failure cases.} Our method tends to fail for some complicated scenarios, e.g., poor lighting, severe occlusions and background interference. Some of such failure cases are shown in Fig.~\ref{fig:failurecase}.


More supplementary materials including demo videos are available at the project website:~\url{https://sites.google.com/site/hemletspose/}. We will make our code publicly available for research uses and also link it at the same project website.


\section{Conclusion}

In this paper, we proposed a simple and highly effective HEMlets-based 3D pose estimation method from a single color image. HEMlets is an easy-to-learn intermediate representation encoding the relative forward-or-backward depth relation for each skeletal part's joints, together with their spatial co-location likelihoods. It is proved very helpful to bridge the input 2D image and the output 3D pose in the learning procedure. We demonstrated the effectiveness of the proposed method tested over the standard benchmarks, yielding a relative accuracy improvement of about 20\% over one best-of-grade method~\cite{sun2017integral} on the Human3.6M benchmark. Good generalization ability is also witnessed for the presented approach. Extending the HEMlets pose estimation network, we further designed a simple parametric 3D human body regression network to estimate the SMPL body shape and pose from the input color image. Extensive experiments have shown the state-of-the-art performance of the proposed HEMlets PoSh method both quantitatively and qualitatively. Specifically, it has achieved the best human body recovery results across different benchmark datasets and evaluation metrics, and notably, obtained the lowest ``surface'' errors for the SURREAL and UP-3D datasets.

We believe the proposed HEMlets idea is actually general, which may potentially benefit other 3D regression problems e.g., scene depth estimation. Future directions also include an optimized real-time system that detects and tracks multiple persons robustly, and reconstructs their 3D poses and body shapes.

\ifCLASSOPTIONcompsoc
  \section*{Acknowledgments}
\else
  \section*{Acknowledgment}
\fi

This work is supported in part by the National Natural Science Foundation of China (Grant No.: 61771201), the Program for Guangdong Introducing Innovative and Enterpreneurial Teams (Grant No.: 2017ZT07X183), the Pearl River Talent Recruitment Program Innovative and Entrepreneurial Teams in 2017 (Grant No.: 2017ZT07X152), the Shenzhen Fundamental Research Fund (Grants No.: KQTD2015033114415450 and ZDSYS201707251409055), and Department of Science and Technology of Guangdong Province Fund (2018B030338001). The authors would like to thank Yulong Shi and Kaiqi Wang for assisting in some early experiments. This work was mainly done when Kun, Nianjuan and Jiangbo were working in Shenzhen Cloudream Technology Co., Ltd.

{\small
\bibliographystyle{ieee_fullname}
\bibliography{HEMlets_PoSh_arXiv}
}


\end{document}